\documentclass[10pt,twocolumn,letterpaper]{article}

\usepackage{iccv}              
\usepackage{enumitem}
\usepackage{xcolor}
\usepackage{amssymb}
\usepackage{graphicx}
\usepackage{pifont}
\usepackage{tcolorbox}

\definecolor{navyblue}{HTML}{0071BC}
\definecolor{hotpink}{HTML}{FF0080}

\definecolor{oai-white}{HTML}{FFFFFF}
\definecolor{oai-black}{HTML}{000000}
\definecolor{oai-red}{HTML}{FF4500}
\definecolor{oai-green}{HTML}{51DA4C}
\definecolor{oai-blue}{HTML}{0000FF}
\definecolor{oai-yellow}{HTML}{FFF639}
\definecolor{oai-magenta}{HTML}{FF45FF}
\definecolor{oai-cyan}{HTML}{00FFFF}
\definecolor{oai-orange}{HTML}{FE7600}
\definecolor{oai-violet}{HTML}{8A2BE2}
\definecolor{oai-brown}{HTML}{A0522D}
\definecolor{oai-green-050}{HTML}{F4FFF4}
\definecolor{oai-green-100}{HTML}{E9FFE8}
\definecolor{oai-green-200}{HTML}{D9FFD8}
\definecolor{oai-green-300}{HTML}{C9FFC7}
\definecolor{oai-green-400}{HTML}{A6FFA3}
\definecolor{oai-green-500}{HTML}{7CF178}
\definecolor{oai-green-600}{HTML}{51DA4C}
\definecolor{oai-green-700}{HTML}{3FA93B}
\definecolor{oai-green-800}{HTML}{2D712A}
\definecolor{oai-green-900}{HTML}{193718}
\definecolor{oai-gray-000}{HTML}{FFFFFF}
\definecolor{oai-gray-100}{HTML}{FAFAFA}
\definecolor{oai-gray-200}{HTML}{F5F5F5}
\definecolor{oai-gray-300}{HTML}{E5E5E5}
\definecolor{oai-gray-400}{HTML}{FFB7A4}
\definecolor{oai-gray-500}{HTML}{CDCDCD}
\definecolor{oai-gray-600}{HTML}{A8A8A8}
\definecolor{oai-gray-700}{HTML}{747474}
\definecolor{oai-gray-800}{HTML}{393939}
\definecolor{oai-gray-900}{HTML}{000000}

\usepackage{xcolor}         
\usepackage{graphicx} 
\usepackage{amssymb} 
\usepackage{booktabs} 
\usepackage{array} 
\usepackage{makecell}
\usepackage{colortbl}
\usepackage{multicol}
\usepackage{multirow}

\definecolor{mygray}{gray}{0.2}
\definecolor{lightred}{rgb}{1, 0.5, 0.5}
\definecolor{lightyellow}{rgb}{1, 0.7, 0.2}

\definecolor{Gray}{gray}{0.9}
\definecolor{mygreen}{rgb}{0.0, 0.5, 0.0}
\definecolor{myred}{rgb}{0.8, 0.25, 0.33}
\definecolor{myblue}{rgb}{0.19, 0.55, 0.91}
\definecolor{uclablue}{rgb}{0.15, 0.45, 0.68}
\definecolor{boxgreen}{rgb}{0.02, 0.66, 0.02}
\definecolor{boxred}{rgb}{0.66, 0.1, 0.1}
\definecolor{boxblue}{rgb}{0.01, 0.01, 0.73}
\definecolor{lightgreen}{rgb}{0.745,0.85,0.68} 
\definecolor{lightorange}{rgb}{0.98,0.76,0.65} 

\usepackage{tcolorbox}
\tcbuselibrary{skins}

\usepackage{xcolor}    
\usepackage{parskip}   

\newcommand{\finding}[2]{
    \vspace{0.05cm}
    \begin{tcolorbox}[
        colback=white!90!gray,     
        colframe=teal!60!black,     
        arc=5pt,                    
        boxsep=5pt,                 
        left=5pt,                  
        right=5pt,                 
        top=4pt,                    
        bottom=4pt,                 
        boxrule=0.8pt,              
        drop shadow=gray!50!white,  
        enhanced jigsaw             
    ]
    \vspace{-0.18cm}
        \paragraph{\textbf{\textit{Finding #1:}}} #2
    \vspace{-0.2cm}
    \end{tcolorbox}
    \vspace{0.05cm}
}

\definecolor{iccvblue}{rgb}{0.21,0.49,0.74}
\usepackage[pagebackref,breaklinks,colorlinks,allcolors=iccvblue]{hyperref}

\def\paperID{****} 
\def\confName{ICCV}
\def\confYear{2025}

\title{
\raisebox{-0.35ex}{\protect\includegraphics[height=2.8\fontcharht\font`\B]{figure/fig0-all-angles-bench-icon.pdf}}
Seeing from Another Perspective:\\ Evaluating Multi-View Understanding in MLLMs}

\author{Chun-Hsiao Yeh\textsuperscript{1*} \quad 
Chenyu Wang\textsuperscript{2*} \quad 
Shengbang Tong\textsuperscript{3} \quad 
Ta-Ying Cheng\textsuperscript{4} \quad 
Ruoyu Wang\textsuperscript{2} \\
Tianzhe Chu\textsuperscript{6} \quad 
Yuexiang Zhai\textsuperscript{1} \quad 
Yubei Chen\textsuperscript{5} \quad 
Shenghua Gao\textsuperscript{2,6} \quad 
Yi Ma\textsuperscript{1,2,6} \quad \\ \\
\textsuperscript{1}UC Berkeley \quad
\textsuperscript{2}TranscEngram \quad
\textsuperscript{3}NYU \quad
\textsuperscript{4}University of Oxford \quad
\textsuperscript{5}UC Davis \quad
\textsuperscript{6}HKU
}

\begin{document}

\twocolumn[{
    \maketitle
    \begin{center}
        \centering
        \captionsetup{type=figure}
        \includegraphics[width=1\linewidth]{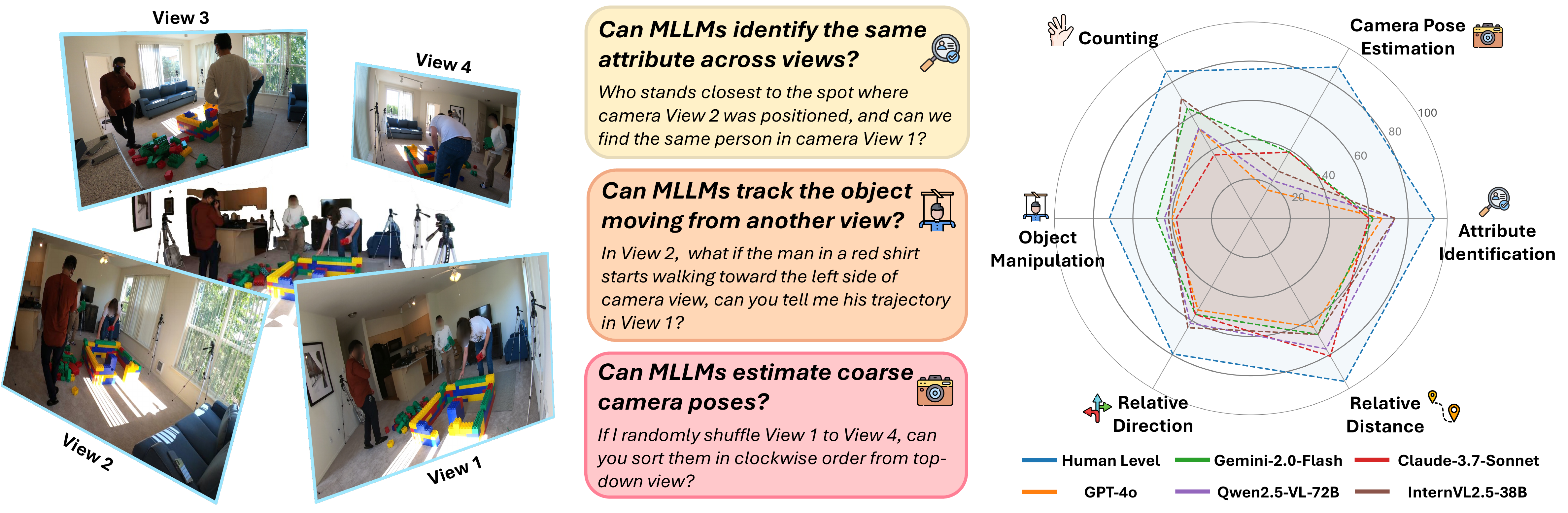}
        \caption{
        We present \textit{All-Angles Bench}, a rich-annotated benchmark with over 2,100 Q\&A pairs from 90 diverse scenes for evaluating multi-view understanding of MLLMs. \textbf{Left and Middle:} An example question setup of multiple views capturing the same scene and the corresponding questions. \textbf{Right:} Accuracies of six notable MLLMs across different question categories.
        }
        \vspace{25pt}
        \label{fig:teaser}
        
    \end{center}
}]

{\let\thefootnote\relax\footnote{$^*$Equal Contribution}}

\begin{abstract}
Multi-view understanding, the ability to  reconcile visual information across diverse viewpoints for effective navigation, manipulation, and 3D scene comprehension, is a fundamental challenge in Multi-Modal Large Language Models (MLLMs) to be used as embodied agents. While recent MLLMs have shown impressive advances in high-level reasoning and planning, they frequently fall short when confronted with multi-view geometric consistency and cross-view correspondence. To comprehensively evaluate the challenges of MLLMs in multi-view scene reasoning, we propose \textit{All-Angles Bench}, a benchmark of over \textbf{2{,}100} human carefully annotated multi-view question--answer pairs across \textbf{90} diverse real-world scenes. Our six tasks (counting, attribute identification, relative distance, relative direction, object manipulation, and camera pose estimation) specifically test model's geometric correspondence and the capacity to align information consistently across views. Our extensive experiments, benchmark on \textbf{27} representative MLLMs including \textit{Gemini-2.0-Flash}, \textit{Claude-3.7-Sonnet}, and \textit{GPT-4o} against human evaluators reveals a substantial performance gap, indicating that current MLLMs remain far from human-level proficiency. Through in-depth analysis, we show that MLLMs are particularly underperforming under two aspects: (1) cross-view correspondence for partially occluded views and (2) establishing the coarse camera poses. These findings highlight the necessity of domain-specific refinements or modules that embed stronger multi-view awareness. We believe that our All-Angles Bench offers valuable insights and contribute to bridging the gap between MLLMs and human-level multi-view understanding. The project and benchmark are publicly available at \url{https://danielchyeh.github.io/All-Angles-Bench/}.
\end{abstract}

\begin{figure*}[!h]
    \centering
    \includegraphics[width=\linewidth]{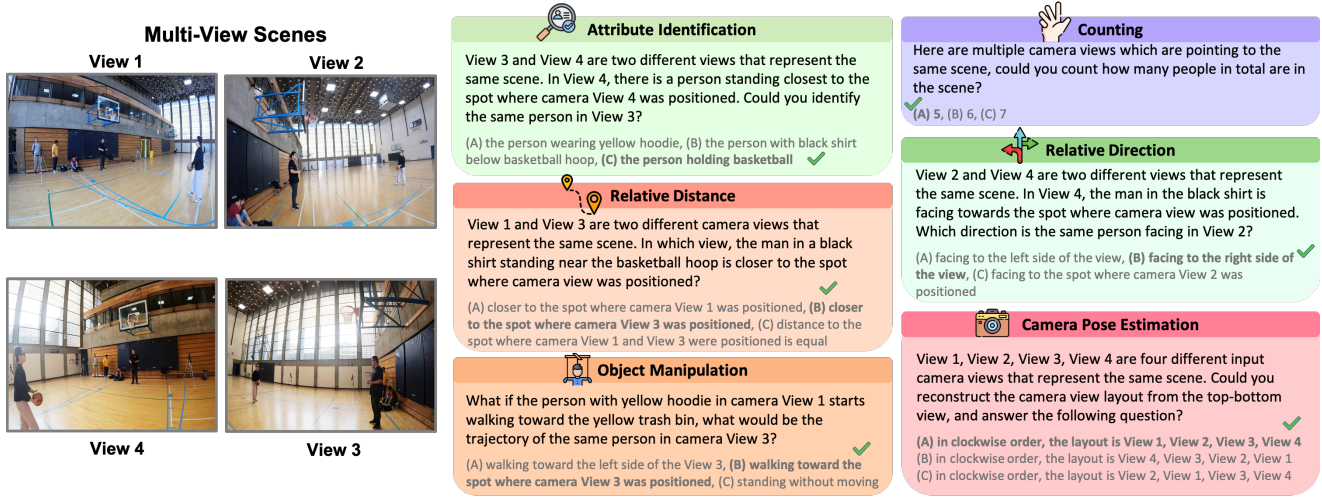}
     \caption{\textbf{Overview of \textit{All-Angles Bench}.} Our benchmark targets a comprehensive view of multi-view understanding, spanning six primary question types. These question types are designed to investigate several major aspects of 3D scene understanding, from creating correspondence between objects to associating relative object and camera poses.} 
    \label{fig:question-type}
\end{figure*}

\section{Introduction}
\label{sec:intro}

Multi-view understanding is a fundamental challenge in bridging machine and human-level understanding~\cite{das2018embodied,yu2019multi,hong20233d} because it underpins an agent’s ability to perceive the environment consistently from diverse viewpoints. By ensuring geometric coherence and cross-view consistency, agents can accurately reconstruct scene layouts and object relationships --- capabilities critical for effective navigation, manipulation, and interaction in the real world~\cite{song2022one,suglia2021embodied}. 
The recent advancement in Multimodal Large Language Models (MLLMs) demonstrates strong capabilities in high-level reasoning and task planning~\cite{li2024llava,hurst2024gpto,Anthropic2024Claude,Gemini,bai2025qwen2,chen2024internvl}, and thus the feasibility of directly using MLLMs as embodied agents is an intriguing research challenge~\cite{huang2022language,driess2023palm,yue2024mmmu,kim2024openvla,niu2024llarva,liu2024moka}. However, such capacities alone are insufficient for generalist embodied agents operating in the real world, where a comprehensive 3D scene understanding and robust multi-view reasoning are pivotal~\cite{jia2024sceneverse, cheng2025spatialrgpt,jatavallabhula2023conceptfusion}. Recent studies survey that MLLMs lacking multi-view scene understanding often commit agent manipulation and navigation errors such as misjudge the target distance, skip partially occluded obstacles --- stemming from limited awareness of multi-view geometry and object relationships~\cite{yu2025inst3dlmminstanceaware3dscene,zhu2024llava}. Since these models must navigate, manipulate, and make decisions in real world environments, it is vital to evaluate (and ultimately strengthen) their multi-view understanding capabilities. Yet, this aspect remains underexplored in details.

To this end, we raise two questions: \emph{(1)~Do MLLMs possess the ability to understand multiple viewpoints simultaneously?} and \emph{(2)~What are the key challenges in MLLMs to gain better multi-view understanding?}

\begin{figure*}[!h]
    \centering
    \includegraphics[width=1\linewidth]{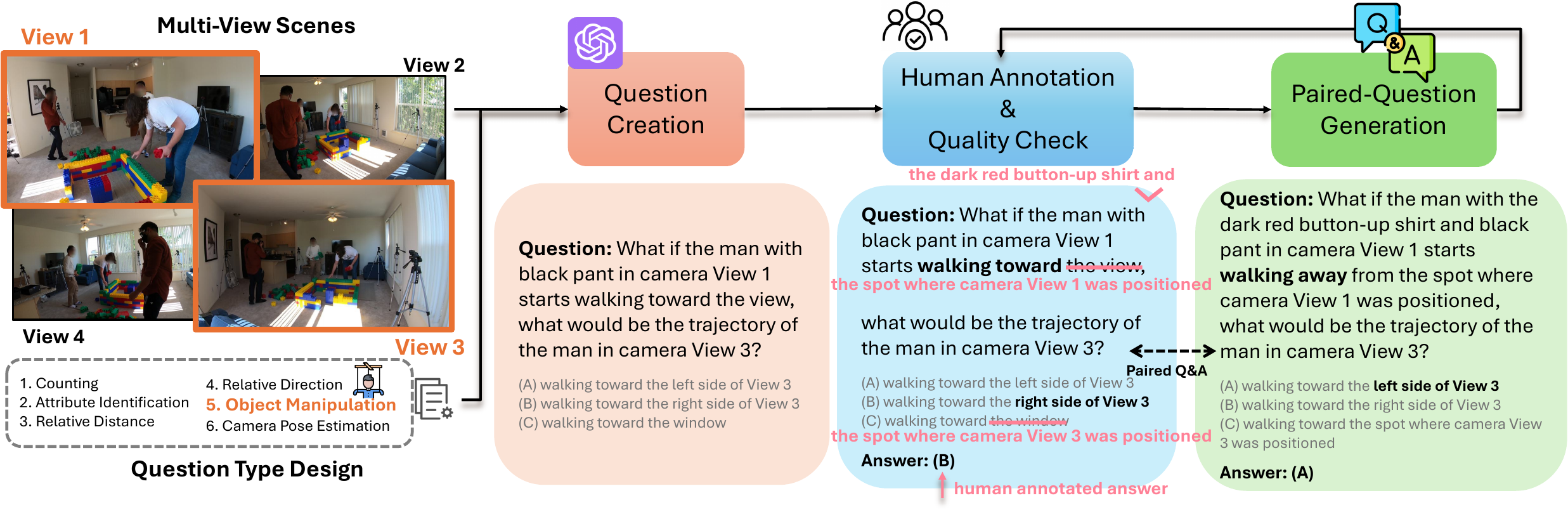}
    \caption{\textbf{\textit{All-Angles Bench} construction pipeline.}
    (1) We collect and curate 90 diverse multi-view scenes and design six tasks that emphasize multi-view reasoning. (2) We generate initial questions via an MLLM, then refine and validate them through \emph{human annotation} to ensure correctness, clarity, and domain relevance. (3) We create paired questions by systematically rephrasing or altering each view perspective while preserving their underlying visual correspondences to evaluate model's cross-view consistency. A final quality-control step removes inconsistent or ambiguous pairs. Note that \emph{counting} and \emph{camera pose estimation} tasks utilize all available views per query, whereas other tasks employ two randomly selected viewpoints.} 
    \label{fig:data-collection}

\end{figure*}

To address these questions and in light of the lack of benchmarks to evaluate multi-view reasoning, we introduce \textit{All-Angles Bench}, comprising over 2{,}100 carefully human-annotated question-answer pairs across 90 diverse multi-view scenes in real world~\cite{grauman2024ego,khirodkar2023ego}. We define six tasks --- \textit{counting, attribute identification, relative distance, relative direction, manipulation, and camera pose estimation} --- with a focus on evaluating MLLM’s geometric understanding and its ability to align information consistently across multi-view scenes. To better evaluate whether models truly possess multi-view capabilities, we also propose a paired question scheme by creating a second question with the same content but with slightly changed wording/order of views. We benchmark 27 representative MLLMs (including Gemini-2.0~\cite{team2023gemini}, Claude-3.7~\cite{Anthropic2024Claude}, and GPT-4o~\cite{OpenAI2024gpt4o}) against human evaluators. As revealed in Figure~\ref{fig:teaser}, a substantial performance gap persists between current MLLMs and human evaluators.

To better understand why MLLMs fall short of human-level multi-view reasoning, we conduct an in-depth analysis of commonly failed questions and tasks, and derive two key findings. \emph{First, MLLMs struggle to identify the same object across multiple views.} We further test whether chain-of-thought prompting --- a technique that has proven effective in other reasoning tasks~\cite{yang2024thinking,zhang2025vlm,rudman2025forgotten} --- could address this limitation. However, our experiments reveal that these linguistic strategies do not provide consistent improvements across models for multi-view reasoning. This suggests that more fundamental domain-specific refinements to multi-view awareness modules or training data are necessary for MLLMs to fully internalize cross-view consistency. \emph{Second, MLLMs often fail to establish correspondence between different viewpoints.} We visualize how models infer scene layouts from multiple perspectives, revealing a consistent inability to accurately estimate camera poses, which in turn impedes performance on tasks like \emph{relative direction} and \emph{object manipulation}. We hope these insights will be helpful to future research towards bringing more better multi-view capabilities in MLLMs.

\section{All-Angles Bench}
\label{sec:method}

The ability to integrate observations of the scene layout from multiple viewpoints is critical for the geometric understanding of MLLMs, which can significantly help with capturing and anticipating interaction outcomes of real-world complex environments safely.

\subsection{Overview of All-Angles Bench}
Most existing benchmarks to evaluate MLLMs primarily rely on single-view or egocentric data, leaving the multi-view consistency and correspondence capabilities of current MLLMs largely unexamined. To address this gap, we introduce \textit{All-Angles Bench}, which comprehensively evaluates MLLMs’ performance across six task categories in multi-view scenarios: \textbf{\textit{(1) Counting:}} Enumerating objects across viewpoints without double-counting or overlooking occluded elements; \textbf{\textit{(2) Attribute Identification:}} Recognizing key properties (e.g., pose, color, shape, orientation) consistently across different viewing perspectives; \textbf{\textit{(3) Relative Distance:}} Estimating object distances when presented with multiple views; \textbf{\textit{(4) Relative Direction:}} Testing the understanding of directional relationships between objects across different views; \textbf{\textit{(5) Object Manipulation:}} Inferring changes in object positions, orientations, or configurations across views; \textbf{\textit{(6) Camera Pose Estimation:}} Evaluating the capacity to estimate viewpoint arrangements or scene layouts from multi-view inputs. Each task addresses a specific dimension of multi-view reasoning, ensuring a thorough assessment of MLLMs’ geometric understanding and their ability to align information across perspectives.

Our \textit{All-Angles Bench} is derived from a curated selection of 90 diverse multi-view scenes sourced from Ego-Exo4D~\cite{grauman2024ego} and EgoHumans~\cite{khirodkar2023ego}, totaling 2{,}132 question–answer pairs. Each question is structured as a multiple-choice questionnaire with three options, only one of which is correct. By collecting multi-view data from varying directions and viewpoints, we generate paired question-answers that highlight differences in appearance, occlusion, and spatial relationships across perspectives while preserving the underlying visual correspondences. The benchmark scenes encompass a wide range of activities (e.g., basketball, soccer, cooking, music playing) and environments (e.g., offices, gym, repair store, kitchen, playground) to ensure broad coverage of real-world scenarios where cross-view reasoning is essential. As shown in Figure~\ref{fig:question-type}, each question targets one of the six task categories outlined above to provide a challenging yet realistic platform for evaluating MLLMs’ geometric understanding and multi-view correspondence.

\subsection{Benchmark Collection Process}
We build a benchmark collection pipeline to effectively generate high quality question-answer pairs for multi-view understanding, as shown in Figure~\ref{fig:data-collection}. To ensure the benchmark quality, all questions were manually annotated by human annotators after collecting and clipping the raw questions.

\noindent\textbf{Data Collection \& Question Type Design.}
We begin by manually selecting 90 diverse multi-view scenes from Ego-Exo4D~\cite{grauman2024ego} and EgoHumans~\cite{khirodkar2023ego}, covering a broad spectrum of activities and environments (e.g., indoor settings, residential areas, industrial spaces) to ensure varied visual contexts. Since the focus of this benchmark is on multi-view analysis, we ensure each scene includes footage captured from at least three viewpoints. We then manually design six task categories spanning fundamental aspects of multi-view understanding: from enumerating and identifying objects across multiple viewpoints (\textit{counting, attribute identification}), to capturing spatial relationships (\textit{relative distance, relative direction}), and analyzing how objects change across views or camera perspectives (\textit{object manipulation, camera pose estimation}). Please see Appendix for further details on the specific question design.

\noindent\textbf{Question Creation \& Human Annotation.}
After collecting our multi-view scenes and designing question templates for each task category, we leverage an MLLM~\cite{OpenAI2024gpt4o} to generate initial questions grounded in the multi-view visual data. Specifically, we generate three questions per category for each multi-view scene except generating one question for \emph{camera pose estimation}. We utilize all available views per query for \emph{counting} and \emph{camera pose estimation} tasks, whereas other tasks employ two randomly selected viewpoints. We hire eight human annotators who carefully examine each question along with the associated multi-view images, removing invalid entries and refining question phrasing in Figure~\ref{fig:data-collection} (middle). This meticulous manual process also involves revising incorrect answer choices and finally annotating the single correct answer. For example, in attribute identification, the MLLM might inconsistently describe an object across two different camera views. In relative direction, it might offer contradictory options --- e.g., \textit{“facing the right side of the camera view”} vs. \textit{“facing the stove”} --- that actually reference the same orientation. Detailed instructions and guidelines for human annotator can be found in the Appendix.

\begin{figure}[!t]
    \centering
    \includegraphics[width=1.0\columnwidth]{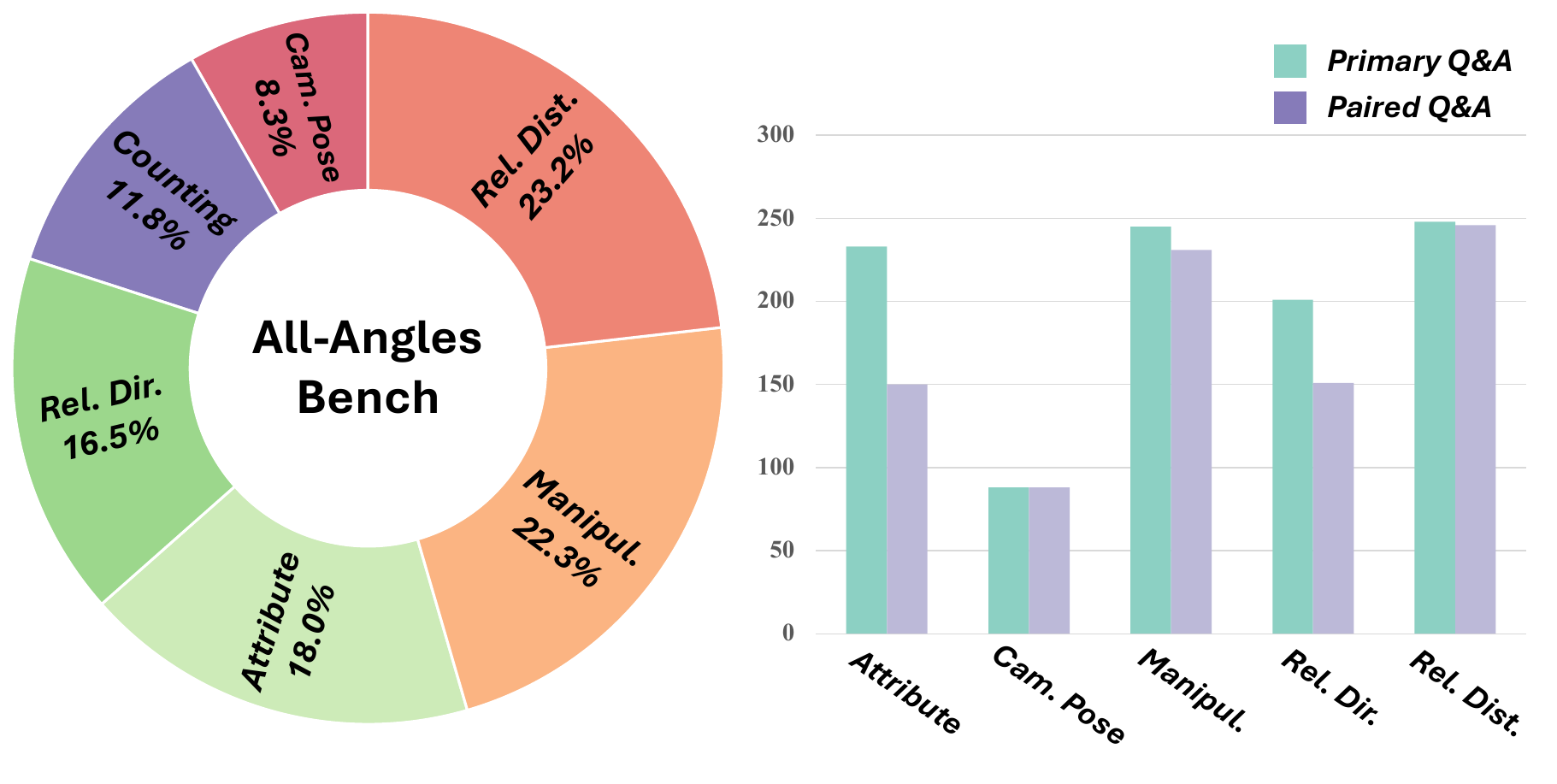}
    \caption{\textbf{Statistical overview of \textit{All-Angles Bench}.} The pie chart shows the distribution of 6 sub-tasks of multi-view understanding. The bar plot illustrates the percentage breakdown by primary and paired question-answers of each sub-task.} 
    \label{fig:bench_stat}
\end{figure}

\noindent\textbf{Paired-Question Generation \& Human Quality Check.}
To rigorously evaluate whether MLLMs truly grasp multi-view concepts, we generate paired questions by systematically rephrasing (i.e., orientation) or altering the original queries (e.g., views) while preserving their underlying visual correspondences and the question structure. For instance, an attribute identification question such as \textit{“Is there a man wearing a yellow hoodie in View 1? Identify him in View 2.”} can be paired with \textit{“There is a man wearing a yellow hoodie in View 2? Identify him in View 1,”} ensuring both questions reference the same individual despite different viewpoint. Likewise, for relative direction, we swap orientations (e.g., left vs. right) and reference views (View 1 vs. View 2). This process is similar to language manipulation in ~\cite{yang2024dynamic,zhu2024dynamic} but requires careful verification of view-to-view consistency. A final human quality check ensures geometric alignment between the paired questions, resulting in 85.3\% of questions having paired counterparts (\textit{counting} task is not involved) --- thereby testing whether MLLMs genuinely understand multi-view scenarios or merely guess answers. The statistics of benchmark is shown in Figure~\ref{fig:bench_stat}.

\section{MLLMs Have Multi-View Understanding?}
\label{sec:exp}

\begin{figure*}[ht!]
    \captionsetup{type=table}
    \vspace{-0.4cm}
    \centering
    \fontsize{10.2pt}{10.0pt}\selectfont
    \setlength\tabcolsep{6pt}  
    \renewcommand{\arraystretch}{1.0}  
    \scalebox{1.0}{
    \begin{tabular}{r|c|cccccccc}
    & & 
    \rotatebox{45}{Attribute} &
    \rotatebox{45}{Cam. Pose} &
    \rotatebox{45}{Counting} & 
    \rotatebox{45}{Manipul.} &
    \rotatebox{45}{Rel. Dir.} &
    \rotatebox{45}{Rel. Dist.}      \\
    Methods & Avg. & \multicolumn{6}{c}{\cellcolor{pink!20}Multiple-Choice Answer} \\
    \hline
    \rowcolor{green!10}
    \multicolumn{1}{l|}{\textcolor{black}{\textit{Performance Against Human (250 Q\&As)}}}  & & & & & & &  \\
    Human Level  & 82.0  & 93.3  & 88.9  & 86.3  & 72.0  & 79.5  & 95.7 \\
    GPT-4o & 52.4  & 66.7  & 16.7  & 52.9  & 40.0  & 53.8  & 63.8  \\ 
    Gemini-2.0-Flash & 58.4  & 62.2  & 38.9  & 64.7  & 48.0  & 56.4  & 68.1 \\
    Claude-3.7-Sonnet & 52.8  & 60.0  & 38.9  & 37.3  & 38.0  & 56.4  & 80.9 \\
    InternVL2.5-38B & 60.8  & 73.3  & 27.8  & 70.6  & 42.0  & 64.1  & 68.1  \\   
    Qwen2.5-VL-72B & 58.4  & 73.3  & 22.2  & 52.9  & 44.0  & 61.5  & 76.6  \\
    \hline
    \rowcolor{green!10}
    \multicolumn{1}{l|}{\textcolor{black}{\textit{Closed-source Models}}}  & & & & & & &  \\
    GPT-4o & 47.8  & 66.8  & \cellcolor{gray!25}{35.8}  & 43.0  & 42.6  & 38.9  & 51.2  \\ 
    Gemini-1.5-Pro & 47.4  & 59.8  & 33.5  & 39.4  & 45.2  & 38.6  & 55.1  \\ 
    Gemini-1.5-Flash & 46.6  & 62.9  & \cellcolor{gray!50}{43.8}  & 35.9  & 43.9  & 33.2  & 52.4  \\ 
    Gemini-2.0-Flash & 52.3  & 68.4  & 33.0  & \cellcolor{gray!50}{64.9}  & 41.0  & 41.8  & 58.9  \\ 
    Claude-3.5-Sonnet & 48.2  & 63.2  & 33.0  & 41.8  & 41.2  & 43.5  & 55.3  \\ 
    Claude-3.7-Sonnet & 50.0  & 68.4  & \cellcolor{gray!25}{35.8}  & 41.4  & 40.1  & 46.9  & 56.7 \\
    \hline
    \rowcolor{green!10}
    \multicolumn{1}{l|}{\textcolor{black}{\textit{Open-source Models}}}  & & & & & & &  \\
    DeepSeek-VL2-Small & 45.5  & 65.3  & 27.8  & 39.0  & 42.6  & 32.7  & 51.6  \\ 
    DeepSeek-VL2 & 47.8  & 70.5  & 24.4  & 39.0  & \cellcolor{gray!50}{46.2}  & 33.5  & 54.7  \\ 
    InternVL2.5-2B & 41.0  & 59.5  & 15.9  & 42.6  & 34.2  & 30.7  & 48.8  \\ 
    InternVL2.5-4B & 45.8  & 66.6  & 18.2  & 47.8  & 36.6  & 35.8  & 54.7  \\ 
    InternVL2.5-8B & 49.9  & 73.9  & 28.4  & 48.6  & 41.6  & 40.3  & 54.5  \\ 
    InternVL2.5-38B & \cellcolor{gray!25}{55.6}  & \cellcolor{gray!50}{80.4}  & 
    31.3  & \cellcolor{gray!25}{56.6}  & 45.2  & 49.7  & 58.7  \\ 
    InternVL2.5-78B & 52.5  & \cellcolor{gray!25}{79.4}  & 27.3  & 52.6  & 39.7  & 43.5  & 59.3  \\ 
    Qwen2.5-VL-3B & 45.2  & 62.7  & 22.2  & 45.0  & 37.2  & 36.4  & 53.8  \\ 
    Qwen2.5-VL-72B & \cellcolor{gray!50}{55.7}  & 77.5  & 29.5  & 55.4  & 43.7  & \cellcolor{gray!50}{54.3}  & \cellcolor{gray!25}{60.7}  \\
    Ovis2-2B & 46.2  & 61.9  & 26.7  & 49.0  & 42.0  & 35.5  & 51.4  \\ 
    Ovis2-4B & 46.6  & 65.5  & 21.6  & 53.4  & 34.0  & 36.1  & 56.9  \\ 
    Ovis2-8B & 49.1  & 70.5  & 17.0  & 49.4  & 43.5  & 41.2  & 54.7  \\ 
    Ovis2-16B & 53.2  & 75.5  & 29.5  & \cellcolor{gray!25}{56.6}  & 44.3  & 46.3  & 56.1  \\ 
    Ovis2-34B & 55.3  & \cellcolor{gray!25}{79.4}  & 26.7  & 53.8  & \cellcolor{gray!50}{46.2}  & \cellcolor{gray!25}{50.6}  & 59.7 \\ 
    Cambrian-8B & 39.2  & 59.8  & 19.9  & 33.1  & 33.0  & 33.0  & 43.5  \\ 
    Cambrian-13B & 36.5  & 59.0  & 25.6  & 30.7  & 27.3  & 32.1  & 37.9  \\ 
    Cambrian-34B & 41.9  & 63.7  & 20.5  & 38.2  & 37.2  & 35.2  & 43.7  \\
    LLaVA-Onevision-Qwen2-7B & 45.9  & 64.5  & 22.2  & 39.4  & 44.5  & 35.2  & 52.0  \\ 
    LLaVA-Onevision-Qwen2-72B & 52.5  & 73.4  & 26.7  & 45.4  & \cellcolor{gray!25}{45.6}  & 46.3  & 60.3  \\
    LLaVA-Video-Qwen2-7B & 42.8  & 64.8  & 12.5  & 42.2  & 32.6  & 37.2  & 50.8  \\ 
    LLaVA-Video-Qwen2-72B & 53.1  & 73.6  & 27.8  & 46.2  & 45.2  & 46.6  & \cellcolor{gray!50}{61.9}  \\
    \hline
    \end{tabular}
    }
     \vspace{-0.2cm}
    \caption{\textbf{Evaluation results for 27 MLLMs.} We consolidate performance from both closed-source and open-source MLLM evaluations. We use \colorbox{gray!50}{deeper-gray} to highlight the top result among all models in each sub-task, while \colorbox{gray!25}{light-gray} marks the second-best result.}
    \label{tab:main_table}
\end{figure*}

\subsection{Evaluation Setup}

\noindent\textbf{Benchmark Models.}
We evaluate a broad spectrum of MLLMs spanning diverse model families, parameter scales, and training paradigms. On the closed-source side, we include three of the most prominent model families --- Gemini-2.0~\cite{team2023gemini}, Claude-3.7~\cite{Anthropic2024Claude}, and GPT-4o~\cite{OpenAI2024gpt4o}. For open-source models, we examine recent breakthroughs from Deepseek-VL2~\cite{wu2024deepseek}, Qwen2.5-VL~\cite{bai2025qwen2}, InternVL2.5~\cite{chen2024expanding}, Cambrian~\cite{tong2024cambrian}, LLaVA-OneVision~\cite{li2024llavaov}, LLaVA-NeXT-Video~\cite{zhang2024llavanextvideo}, and OVIS~\cite{lu2024ovis}. In all experiments, we follow standard protocols and set the temperature to zero unless otherwise specified.

\noindent\textbf{Human Evaluation.} We randomly select a subset of 250 questions from our \textit{All-Angles Bench} --- encompassing all six task categories for evaluation by human annotators, each of whom independently answers every question. For fair comparison, we also report performance of Gemini-2.0-Flash, Claude-3.7-Sonnet, GPT-4o, Qwen2.5-VL-72B, and InternVL2.5-38B on this subset.

\subsection{Results}
As the primary results shown in Table~\ref{tab:main_table}, there remains a substantial performance gap between both of closed- and open-source MLLMs and human-level multi-view understanding. We post several findings we observe.

\finding{1}{\textit{Simple task for human like coarse camera pose estimation poses challenges for MLLMs.}}

While humans approaching near-perfect accuracy on multiple tasks in our \textit{All-Angles Bench}, both open- and closed-source models often struggle. For example, in camera pose estimation, human annotators achieve 88.9\% accuracy when ordering multiple camera perspectives, whereas state-of-the-art MLLMs such as Gemini-2.0-Flash, Qwen2.5-VL-72B, and InternVL2.5-38B trail behind \textbf{over 50\% margins}. Many open-source MLLMs perform even worse than random guessing. Common errors include failures to reconcile viewpoint transitions and misinterpretations of geometric relationships, underscoring the persistent gap between human-level capabilities and current MLLM performance.

\finding{2}{\textit{Certain open-source MLLMs surpass closed-source ones in orientation-sensitive tasks.}}

Interestingly, Ovis2-34B~\cite{lu2024ovis} and Qwen2.5‐VL‐72B~\cite{bai2025qwen2} outperform leading closed-source models such as Gemini-2.0~\cite{team2023gemini} and Claude-3.7-Sonnet~\cite{Anthropic2024Claude} on \textit{object manipulation} and \textit{relative direction}. We observe that Qwen2.5-VL-72B integrates robust video understanding and fine-grained visual grounding modules (as highlighted in its model report), positioning it well to capture how objects re-orient across different viewpoints. The specialized, video-focused training regimes observed in these open-source models, which emphasize frame-by-frame orientation tracking and spatial grounding --- crucial for handling multi-view scenes. While it is unclear whether closed-source models train with similar strategies, this findings can still be a good indicator that domain-specific refinement can yield better performances in tasks tackling orientation and geometric reasoning.

\begin{figure*}[!t]
    \centering
    \includegraphics[width=1\linewidth]{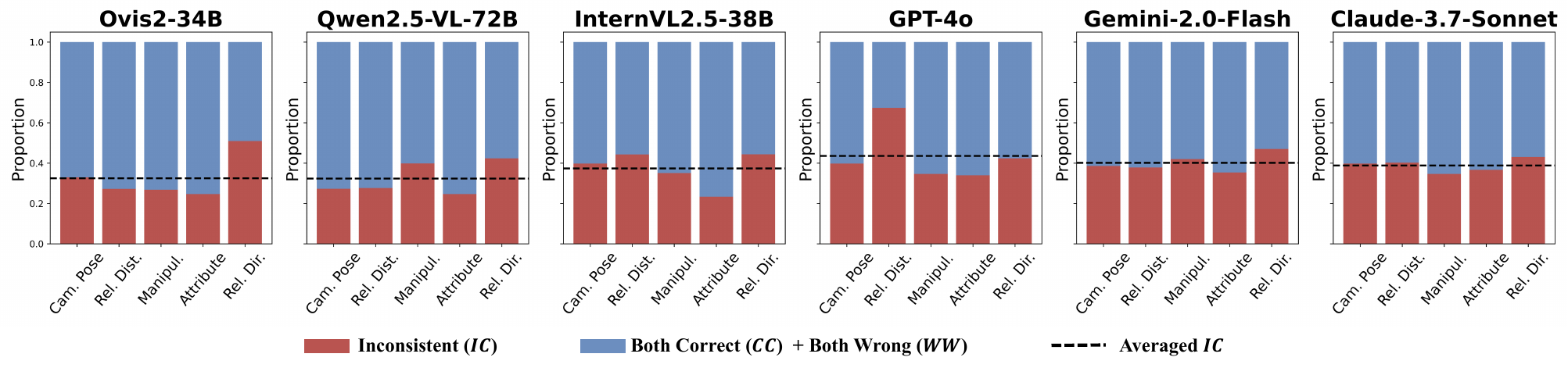}
    \caption{\textbf{Paired question-answers inconsistency across 6 MLLMs.} We report the proportions of \textit{\textbf{IC}} and \textit{\textbf{CC}} + \textit{\textbf{WW}}. Notably, GPT-4o struggles with relative distance (around 70\% inconsistency). Gemini-2.0-Flash and Claude-3.7-Sonnet exhibit more balanced performance, whereas Ovis2-34B and GPT-4o vary considerably across tasks.} 
    \label{fig:pair_data}
\end{figure*}

\begin{figure}[!t]
    \centering
    \includegraphics[width=1\columnwidth]{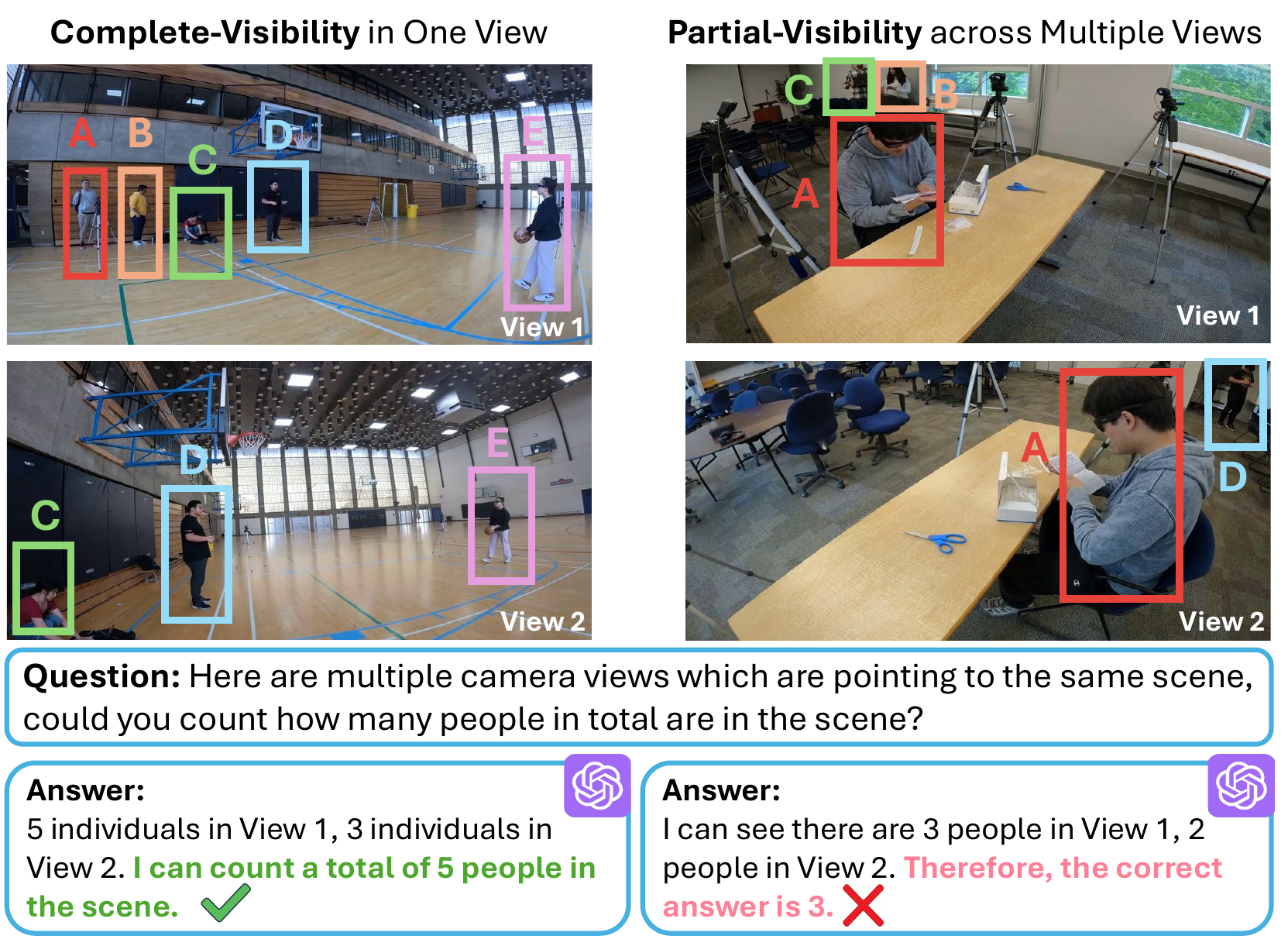}
    \caption{\textbf{Complete- and Partial-visibility counting.} While MLLMs often succeed when everyone is visible in one viewpoint, they sometimes fail to reconcile fragmented information across views, as shown by GPT‐4o occasionally picks the largest per‐view count rather than reconciling individuals across views.} 
    \label{fig:counting-gpt4}
\end{figure}

\subsection{MLLM's Robustness on Paired Questions}
While the correctness of one single question indicates how often a model answers an isolated question correctly, it does not capture whether the model remains consistent when presented with semantically equivalent queries from different viewpoints or rephrasings. To investigate this, we also propose to look into the proportions of questions where the answers are inconsistent with one another. 

First, we classify each paired instance into three scenarios: 1) \textbf{\textit{CC (Both Correct)}} when the model answers both the primary and paired question correctly, 2) \textbf{\textit{WW (Both Wrong)}} when it fails both versions, 
and 3) \textbf{\textit{IC (Inconsistent)}} when the model answers one version correctly but fails the other. 
We are particularly interested in the case of \textbf{\textit{IC}}, as this shows the number of questions where the model answered correctly but does not in fact reflect correct multi-view understanding, as simply changing the order or rephrasing the question leads to a wrong answer.

As shown in Figure~\ref{fig:pair_data}, we report the proportions of \textit{\textbf{IC}} (inconsistent) outcomes across six leading MLLMs --- three open-source (Ovis2-34B, Qwen2.5-VL-72B, InternVL2.5-38B) and three closed-source (GPT-4o, Gemini-2.0-Flash, Claude-3.7-Sonnet). We have several observations: 1) GPT-4o exhibits notably high inconsistency score \textbf{\textit{IC}} (around 70\%) on relative distance tasks, whereas the other five models generally have around 40\% inconsistency in this category, 2) All models struggle with relative direction; all surpasses 40\% inconsistency \textbf{\textit{IC}}, highlighting the challenge of reasoning about orientation shifts in multi-view scenarios, 3) Gemini-2.0-Flash and Claude-3.7-Sonnet remain fairly balanced inconsistency across overall question types, while Ovis2-34B and GPT-4o vary significantly across tasks.

\section{Why Do MLLMs Struggle with Multi-View Understanding?}
\label{sec:analysis}

To investigate specific weaknesses of MLLMs in multi-view comprehension, we evaluate each question type in our \textit{All-Angles Bench}. We select the top-performing closed-source and open-source MLLMs in our benchmark and systematically identify where these models succeed or fail in understanding multi-view scenarios.

\subsection{Failure of Multi-View Correspondence}
\label{sec:corre-fail}

We first investigate the multi-view counting task since we are curious about the discrepancy between egocentric view and multi-view counting. We begin our analysis by examining counting questions especially counting on \textit{how many people in total are in the scene}. We find that MLLMs typically succeed in the \textbf{complete-visibility in one view} scenario (i.e., when all individuals are visible within a single view), but frequently fail in \textbf{the partial-visibility across multiple views} scenario when partial information is distributed across multiple viewpoints (e.g., Person A and B in View 1, and Person C and D in View 2). As illustrated in Figure~\ref{fig:counting-gpt4}, GPT-4o occasionally handles these scenarios by simply counting the number of people per view and choosing the highest count, neglecting to reconcile individuals across different perspectives and thus leading to errors.

\noindent\textbf{Can Reasoning Injection Improve MLLM's Ability?} To investigate whether linguistic reasoning can enhance MLLMs' multi-view understanding, we randomly select 55 scenes from our 90-scene \textit{All-Angles Bench}, excluding those with only a single person or with insufficient partial-visibility. In each chosen scene, all individuals are visible in at least one camera view (see Figure~\ref{fig:counting-gpt4}, left). We then create a \textit{paired} version of these scenes by manually cropping footage so that key information is split across multiple viewpoints (e.g., Person A and B in View 1, and Person C and D in View 2). This setup enables a fair comparison of MLLMs' performance under the same set of complete-visibility versus partial-visibility conditions.

Prompting techniques have shown promise in enhancing the reasoning and problem-solving capabilities of large models across diverse tasks. Motivated by these findings, we explore whether such linguistic prompts can also bolster the visual-spatial proficiency of MLLMs in multi-view settings. Specifically, we introduce an \textit{Identification CoT} strategy, which instructs the model to (1) provide a detailed description of each visible individual --- noting appearance, clothing, orientation, and interactions with nearby people or objects, (2) cross-reference these descriptions across all views to avoid double-counting, and (3) provide a final tally of unique entities. The detailed prompt of \textit{Identification CoT} could be found in Appendix. We also report two additional CoT strategies, \textit{Zero-Shot CoT} and \textit{Self Consistency /w CoT} which were used in~\cite{yang2024thinking}, for comparison.

\begin{figure}[!t]
    \centering
    \includegraphics[width=1\columnwidth]{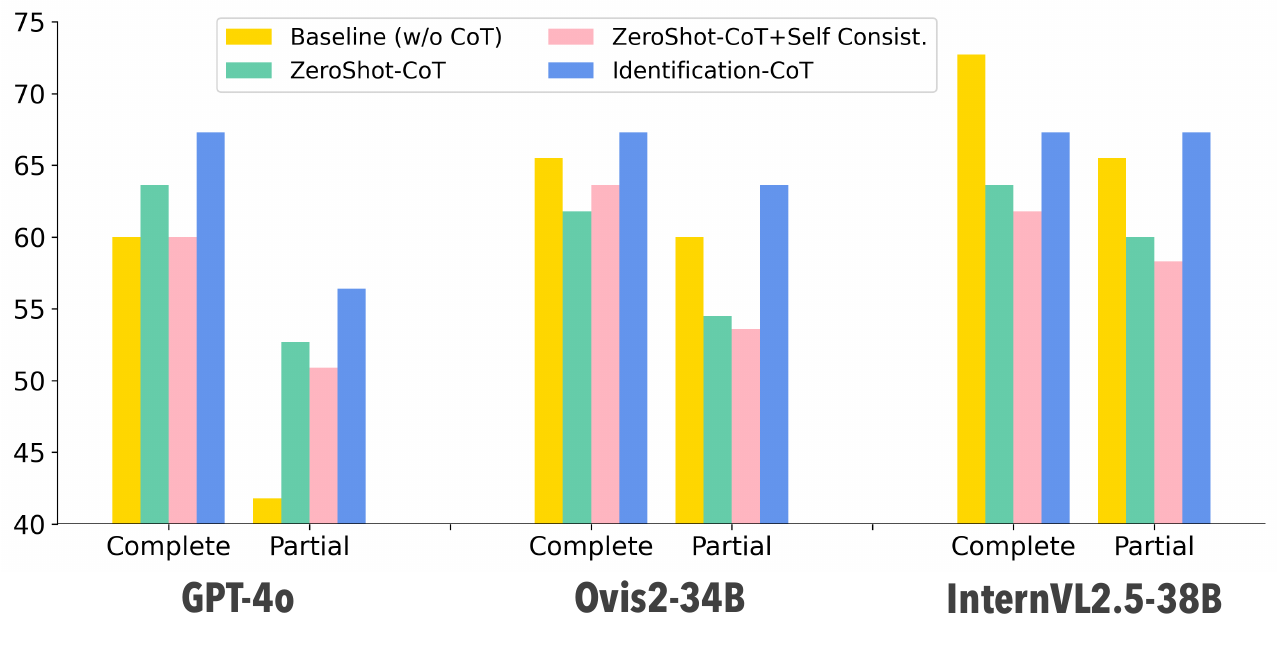}
    \caption{\textbf{Analysis of reasoning prompt strategies.} We report the effectiveness of \textit{Zero-Shot CoT}, \textit{Self-Consistency}, and \textit{Identification CoT} --- across GPT-4o, Ovis2-34B, and InternVL2.5-38B under complete-view and partial-view settings. While CoT variations delivers notable gains in partial-visibility scenarios in GPT-4o, its impact diminishes for models already be robust at multi-view counting (e.g., InternVL2.5-38B). These results indicate that refining reasoning prompt alone is insufficient; specialized multi-view training may be necessary to excel on \textit{All-Angles Bench}.} 
    \label{fig:gpt4-cot}
\end{figure}

We evaluate three prompting strategies --- \textit{Zero-Shot CoT}, \textit{Self-Consistency}, and \textit{Identification CoT} --- across three leading MLLMs: GPT-4o~\cite{OpenAI2024gpt4o}, Ovis2-34B~\cite{lu2024ovis}, and InternVL2.5-38B~\cite{chen2024internvl} chosen for their varying levels of counting proficiency. As GPT-4o's results shown in Figure~\ref{fig:gpt4-cot}, both \textit{Zero-Shot CoT} and \textit{Self-Consistency} yield relative gains of approximately 15\% each over the no-prompting baseline. Notably, \textit{Identification CoT} provides a substantial improvement under partial-visibility conditions, suggesting that explicit entity descriptions and cross-view consistency checks are pivotal for accurate reasoning when some individuals or objects are only partially visible across different views. However, when the model already possesses robust multi-view counting capabilities (e.g., InternVL2.5-38B), the benefits of additional prompting diminish and can even degrade performance, as observed with InternVL2.5-38B. This phenomenon echoes findings in~\cite{yang2024thinking}, where CoT methods offered limited advantages for strong spatial-reasoning models such as Gemini-1.5~\cite{team2023gemini}. We hypothesize that, beyond these prompt reasoning strategies, architectures or training methods specialized for multi-view scenarios --- incorporating domain-specific data or spatial-aware modules may be necessary to further advance MLLMs' performance, rather than relying solely on enhanced prompt engineering.

\begin{figure*}[!h]
    \centering
    \includegraphics[width=\linewidth]{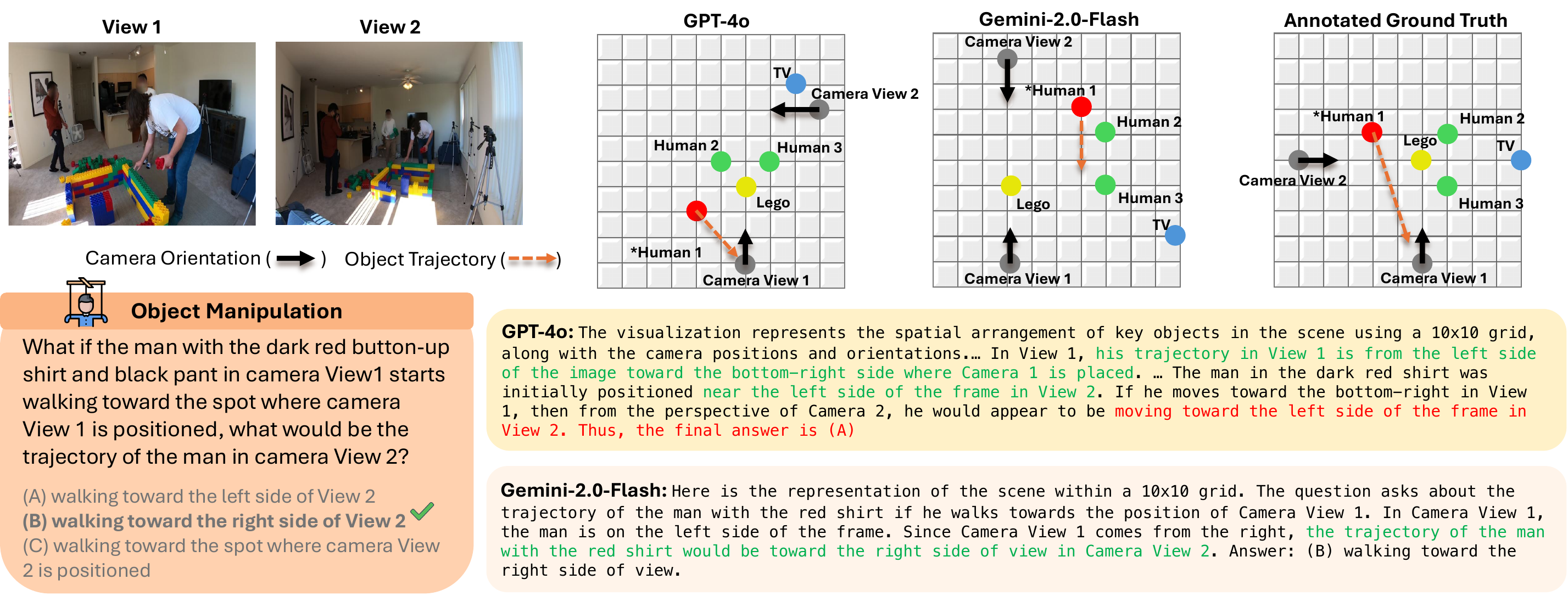}
    \caption{\textbf{Visualization of multi-view scene reconstruction and camera pose alignment.}
    Although GPT-4o and Gemini-2.0-Flash both display moderate proficiency in single-view scene reconstruction, they struggle when aligning two different camera perspectives. Misidentifying camera poses leads to incorrect directional reasoning --- such as tracking a person’s trajectory from View~1 to View~2 which needs multi-view consistency in current MLLMs. }
    \label{fig:scene-est}
\end{figure*}

\begin{figure}[!t]
    \centering
    \includegraphics[width=1\columnwidth]{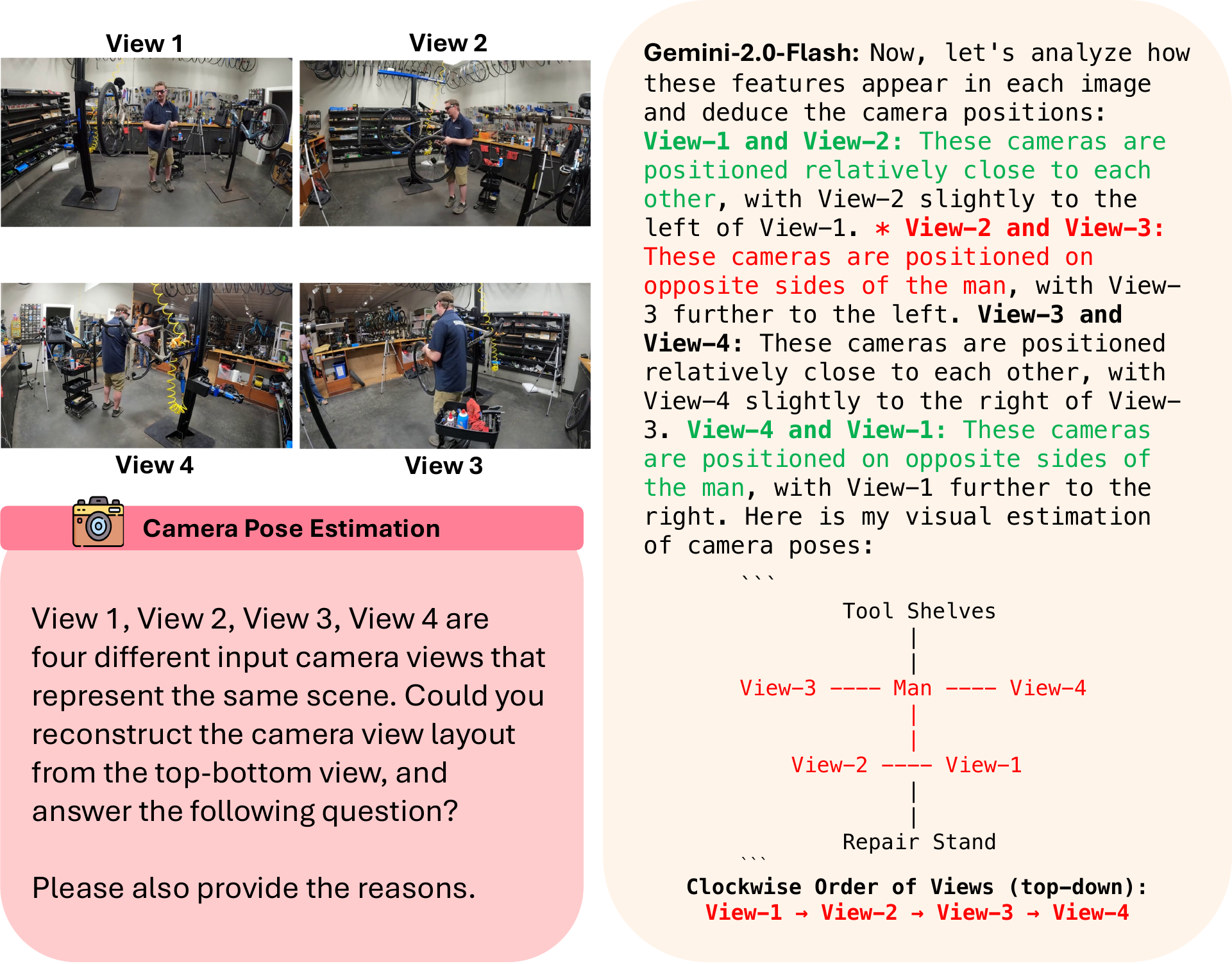}
    \caption{\textbf{Visualization of camera pose estimation.}  When asked to order the camera poses in clockwise order, MLLMs fail completely despite providing detailed reasonings.} 
    \label{fig:cam-pose-est}
\end{figure}

\subsection{Failure with Coarse Camera Estimation}

We also observe that MLLMs often struggle with \emph{orientation-sensitive} challenges (as highlighted in Table~\ref{tab:main_table}), such as estimating camera poses, object directions, and tracking object trajectories across multiple viewpoints. To investigate how these shortcomings manifest, we design a visualization prompt inspired by~\cite{yang2024thinking}, wherein each object’s center is mapped to a $10\times10$ grid and camera view poses are depicted as dot with directional arrow. Specifically, we task GPT-4o~\cite{hurst2024gpto} and Gemini-2.0-Flash~\cite{team2023gemini} with inferring both object and camera's positions and orientations from multi-view images (see Appendix for prompt details).

As illustrated in Figure~\ref{fig:scene-est} (\textit{object manipulation}) and Figure~\ref{fig:cam-pose-est} (\textit{camera pose estimation}), many orientation-related errors stem from the model’s inability to reconcile viewpoint transformations. Instead of maintaining consistent object correspondences, the model frequently misaligns camera coordinates or overlooks background cues critical for geometric reasoning. This not only impacts camera pose estimation but also complicates downstream tasks such as \emph{relative direction} or \emph{object manipulation} where fine-grained rotational and positional cues are essential. These observations echo the findings from Section~\ref{sec:corre-fail}, suggesting that domain-specific training or architectural refinements, especially those emphasizing viewpoint consistency may necessary to close the gap in multi-view understanding.

\section{Related Works}
\label{sec:related-work}

\noindent\textbf{Multimodal Large Language Models.} 
MLLMs~\citep{liu2023visual, liu2023improved, dai2024instructblip, tong2024cambrian, bai2023qwen, li2024llavaov, hurst2024gpto, Gemini} have demonstrated impressive capabilities across various tasks~\citep{masry2022chartqa, yue2023mmmu, liu2023mmbench, lu2023mathvista, tong2024eyes, yue2024mmmu, li2024seed} and applications~\citep{zhai2024fine, zhou2024transfusion, tong2024metamorph, bai2025digirl, chu2025sft, xu2024cad, wang2024mllm}. In particular, an increasing number of studies~\citep{hong20233d, driess2023palm, chen2024spatialvlm, chen2024longvila} are focusing on video understanding and, more broadly, exploring model capabilities for embodied real-world tasks. Our work contributes to this growing area by: 1) providing a timely evaluation benchmark for assessing current and future models' abilities in multi-view perception --- a fundamental capability for 3D and 4D tasks; and 2) offering an analysis of why current models struggle with multi-view understanding.

\noindent\textbf{Benchmarking Visual Spatial Ability.}
There are recently more works~\citep{mangalam2023egoschema, fu2024video, yang2024thinking, li2024mvbench, chandrasegaran2025hourvideo, li2025embodied} studying video and visual spatial ability of MLLMs. Our work is most relevant to VideoMME~\citep{fu2024video}, VSI-Bench~\citep{yang2024thinking} and MV-Bench~\citep{li2024mvbench}. VideoMME extensively evaluates video understanding but emphasizes temporal reasoning~\citep{fu2024video}. VSI-Bench specifically targets spatial intelligence through egocentric video scenarios~\citep{yang2024thinking}. MV-Bench also focuses on comprehensive multi-modal temporal understanding~\citep{li2024mvbench}.  Our work focuses on multi-view understanding, a cornerstone for robust 3D and 4D reasoning in MLLMs. Unlike previous work that primarily assess single-view or temporal reasoning, we explicitly evaluate how models align geometric and semantic information across multiple viewpoints. We further provide a detailed breakdown analysis that dissects model deficiencies in multi-view understanding.

\section{Conclusion}

In conclusion, we introduce \textit{All-Angles Bench}, a comprehensive benchmark to evaluate MLLMs' multi-view understanding. Our evaluation of 27 representative models across over 2,100 annotated multi-view question-answer pairs in the six tasks, we reveal significant limitations in geometric consistency and cross-view correspondence, particularly in cross-view identification and camera pose estimation. These findings highlight the need for domain-specific training to enhance MLLMs' multi-view reasoning, providing insights toward achieving human-level performance.

{
    \small
    \bibliographystyle{ieeenat_fullname}
    \bibliography{main}

\begin{thebibliography}{66}
\providecommand{\natexlab}[1]{#1}
\providecommand{\url}[1]{\texttt{#1}}
\expandafter\ifx\csname urlstyle\endcsname\relax
  \providecommand{\doi}[1]{doi: #1}\else
  \providecommand{\doi}{doi: \begingroup \urlstyle{rm}\Url}\fi

\bibitem[Anthropic(2024)]{Anthropic2024Claude}
Anthropic.
\newblock Claude, 2024.

\bibitem[Bai et~al.(2025{\natexlab{a}})Bai, Zhou, Pan, Cemri, Suhr, Levine, and Kumar]{bai2025digirl}
Hao Bai, Yifei Zhou, Jiayi Pan, Mert Cemri, Alane Suhr, Sergey Levine, and Aviral Kumar.
\newblock Digirl: Training in-the-wild device-control agents with autonomous reinforcement learning.
\newblock \emph{Advances in Neural Information Processing Systems}, 37:\penalty0 12461--12495, 2025{\natexlab{a}}.

\bibitem[Bai et~al.(2023)Bai, Bai, Yang, Wang, Tan, Wang, Lin, Zhou, and Zhou]{bai2023qwen}
Jinze Bai, Shuai Bai, Shusheng Yang, Shijie Wang, Sinan Tan, Peng Wang, Junyang Lin, Chang Zhou, and Jingren Zhou.
\newblock Qwen-vl: A versatile vision-language model for understanding, localization, text reading, and beyond.
\newblock 2023.

\bibitem[Bai et~al.(2025{\natexlab{b}})Bai, Chen, Liu, Wang, Ge, Song, Dang, Wang, Wang, Tang, et~al.]{bai2025qwen2}
Shuai Bai, Keqin Chen, Xuejing Liu, Jialin Wang, Wenbin Ge, Sibo Song, Kai Dang, Peng Wang, Shijie Wang, Jun Tang, et~al.
\newblock Qwen2. 5-vl technical report.
\newblock \emph{arXiv preprint arXiv:2502.13923}, 2025{\natexlab{b}}.

\bibitem[Chandrasegaran et~al.(2025)Chandrasegaran, Gupta, Hadzic, Kota, He, Eyzaguirre, Durante, Li, Wu, and Li]{chandrasegaran2025hourvideo}
Keshigeyan Chandrasegaran, Agrim Gupta, Lea~M Hadzic, Taran Kota, Jimming He, Crist{\'o}bal Eyzaguirre, Zane Durante, Manling Li, Jiajun Wu, and Fei-Fei Li.
\newblock Hourvideo: 1-hour video-language understanding.
\newblock \emph{Advances in Neural Information Processing Systems}, 37:\penalty0 53168--53197, 2025.

\bibitem[Chen et~al.(2024{\natexlab{a}})Chen, Xu, Kirmani, Ichter, Sadigh, Guibas, and Xia]{chen2024spatialvlm}
Boyuan Chen, Zhuo Xu, Sean Kirmani, Brain Ichter, Dorsa Sadigh, Leonidas Guibas, and Fei Xia.
\newblock Spatialvlm: Endowing vision-language models with spatial reasoning capabilities.
\newblock In \emph{Proceedings of the IEEE/CVF Conference on Computer Vision and Pattern Recognition}, pages 14455--14465, 2024{\natexlab{a}}.

\bibitem[Chen et~al.(2024{\natexlab{b}})Chen, Xue, Li, Hu, Zhu, Li, Fang, Tang, Yang, Liu, et~al.]{chen2024longvila}
Yukang Chen, Fuzhao Xue, Dacheng Li, Qinghao Hu, Ligeng Zhu, Xiuyu Li, Yunhao Fang, Haotian Tang, Shang Yang, Zhijian Liu, et~al.
\newblock Longvila: Scaling long-context visual language models for long videos.
\newblock \emph{arXiv preprint arXiv:2408.10188}, 2024{\natexlab{b}}.

\bibitem[Chen et~al.(2024{\natexlab{c}})Chen, Wang, Cao, Liu, Gao, Cui, Zhu, Ye, Tian, Liu, et~al.]{chen2024expanding}
Zhe Chen, Weiyun Wang, Yue Cao, Yangzhou Liu, Zhangwei Gao, Erfei Cui, Jinguo Zhu, Shenglong Ye, Hao Tian, Zhaoyang Liu, et~al.
\newblock Expanding performance boundaries of open-source multimodal models with model, data, and test-time scaling.
\newblock \emph{arXiv preprint arXiv:2412.05271}, 2024{\natexlab{c}}.

\bibitem[Chen et~al.(2024{\natexlab{d}})Chen, Wu, Wang, Su, Chen, Xing, Zhong, Zhang, Zhu, Lu, et~al.]{chen2024internvl}
Zhe Chen, Jiannan Wu, Wenhai Wang, Weijie Su, Guo Chen, Sen Xing, Muyan Zhong, Qinglong Zhang, Xizhou Zhu, Lewei Lu, et~al.
\newblock Internvl: Scaling up vision foundation models and aligning for generic visual-linguistic tasks.
\newblock In \emph{Proceedings of the IEEE/CVF Conference on Computer Vision and Pattern Recognition}, pages 24185--24198, 2024{\natexlab{d}}.

\bibitem[Cheng et~al.(2025)Cheng, Yin, Fu, Guo, Yang, Kautz, Wang, and Liu]{cheng2025spatialrgpt}
An-Chieh Cheng, Hongxu Yin, Yang Fu, Qiushan Guo, Ruihan Yang, Jan Kautz, Xiaolong Wang, and Sifei Liu.
\newblock Spatialrgpt: Grounded spatial reasoning in vision-language models.
\newblock \emph{Advances in Neural Information Processing Systems}, 37:\penalty0 135062--135093, 2025.

\bibitem[Chu et~al.(2025)Chu, Zhai, Yang, Tong, Xie, Schuurmans, Le, Levine, and Ma]{chu2025sft}
Tianzhe Chu, Yuexiang Zhai, Jihan Yang, Shengbang Tong, Saining Xie, Dale Schuurmans, Quoc~V Le, Sergey Levine, and Yi Ma.
\newblock Sft memorizes, rl generalizes: A comparative study of foundation model post-training.
\newblock \emph{arXiv preprint arXiv:2501.17161}, 2025.

\bibitem[Dai et~al.(2024)Dai, Li, Li, Tiong, Zhao, Wang, Li, Fung, and Hoi]{dai2024instructblip}
Wenliang Dai, Junnan Li, Dongxu Li, Anthony Meng~Huat Tiong, Junqi Zhao, Weisheng Wang, Boyang Li, Pascale~N Fung, and Steven Hoi.
\newblock Instructblip: Towards general-purpose vision-language models with instruction tuning.
\newblock In \emph{NeurIPS}, 2024.

\bibitem[Das et~al.(2018)Das, Datta, Gkioxari, Lee, Parikh, and Batra]{das2018embodied}
Abhishek Das, Samyak Datta, Georgia Gkioxari, Stefan Lee, Devi Parikh, and Dhruv Batra.
\newblock Embodied question answering.
\newblock In \emph{Proceedings of the IEEE conference on computer vision and pattern recognition}, pages 1--10, 2018.

\bibitem[Driess et~al.(2023)Driess, Xia, Sajjadi, Lynch, Chowdhery, Wahid, Tompson, Vuong, Yu, Huang, et~al.]{driess2023palm}
Danny Driess, Fei Xia, Mehdi~SM Sajjadi, Corey Lynch, Aakanksha Chowdhery, Ayzaan Wahid, Jonathan Tompson, Quan Vuong, Tianhe Yu, Wenlong Huang, et~al.
\newblock Palm-e: An embodied multimodal language model.
\newblock 2023.

\bibitem[Duan et~al.(2024)Duan, Yang, Qiao, Fang, Chen, Liu, Dong, Zang, Zhang, Wang, et~al.]{duan2024vlmevalkit}
Haodong Duan, Junming Yang, Yuxuan Qiao, Xinyu Fang, Lin Chen, Yuan Liu, Xiaoyi Dong, Yuhang Zang, Pan Zhang, Jiaqi Wang, et~al.
\newblock Vlmevalkit: An open-source toolkit for evaluating large multi-modality models.
\newblock In \emph{Proceedings of the 32nd ACM international conference on multimedia}, pages 11198--11201, 2024.

\bibitem[Fu et~al.(2024)Fu, Dai, Luo, Li, Ren, Zhang, Wang, Zhou, Shen, Zhang, et~al.]{fu2024video}
Chaoyou Fu, Yuhan Dai, Yongdong Luo, Lei Li, Shuhuai Ren, Renrui Zhang, Zihan Wang, Chenyu Zhou, Yunhang Shen, Mengdan Zhang, et~al.
\newblock Video-mme: The first-ever comprehensive evaluation benchmark of multi-modal llms in video analysis.
\newblock \emph{arXiv preprint arXiv:2405.21075}, 2024.

\bibitem[Google(2023)]{Gemini}
Google.
\newblock Gemini, 2023.

\bibitem[Grauman et~al.(2024)Grauman, Westbury, Torresani, Kitani, Malik, Afouras, Ashutosh, Baiyya, Bansal, Boote, et~al.]{grauman2024ego}
Kristen Grauman, Andrew Westbury, Lorenzo Torresani, Kris Kitani, Jitendra Malik, Triantafyllos Afouras, Kumar Ashutosh, Vijay Baiyya, Siddhant Bansal, Bikram Boote, et~al.
\newblock Ego-exo4d: Understanding skilled human activity from first-and third-person perspectives.
\newblock In \emph{Proceedings of the IEEE/CVF Conference on Computer Vision and Pattern Recognition}, pages 19383--19400, 2024.

\bibitem[Hong et~al.(2023)Hong, Lin, Du, Chen, Tenenbaum, and Gan]{hong20233d}
Yining Hong, Chunru Lin, Yilun Du, Zhenfang Chen, Joshua~B Tenenbaum, and Chuang Gan.
\newblock 3d concept learning and reasoning from multi-view images.
\newblock In \emph{Proceedings of the IEEE/CVF Conference on Computer Vision and Pattern Recognition}, pages 9202--9212, 2023.

\bibitem[Huang et~al.(2022)Huang, Abbeel, Pathak, and Mordatch]{huang2022language}
Wenlong Huang, Pieter Abbeel, Deepak Pathak, and Igor Mordatch.
\newblock Language models as zero-shot planners: Extracting actionable knowledge for embodied agents.
\newblock In \emph{International conference on machine learning}, pages 9118--9147. PMLR, 2022.

\bibitem[Hurst et~al.(2024)Hurst, Lerer, Goucher, Perelman, Ramesh, Clark, Ostrow, Welihinda, Hayes, Radford, et~al.]{hurst2024gpto}
Aaron Hurst, Adam Lerer, Adam~P Goucher, Adam Perelman, Aditya Ramesh, Aidan Clark, AJ Ostrow, Akila Welihinda, Alan Hayes, Alec Radford, et~al.
\newblock Gpt-4o system card.
\newblock \emph{arXiv preprint arXiv:2410.21276}, 2024.

\bibitem[Jatavallabhula et~al.(2023)Jatavallabhula, Kuwajerwala, Gu, Omama, Chen, Maalouf, Li, Iyer, Saryazdi, Keetha, et~al.]{jatavallabhula2023conceptfusion}
Krishna~Murthy Jatavallabhula, Alihusein Kuwajerwala, Qiao Gu, Mohd Omama, Tao Chen, Alaa Maalouf, Shuang Li, Ganesh Iyer, Soroush Saryazdi, Nikhil Keetha, et~al.
\newblock Conceptfusion: Open-set multimodal 3d mapping.
\newblock \emph{arXiv preprint arXiv:2302.07241}, 2023.

\bibitem[Jia et~al.(2024)Jia, Chen, Yu, Wang, Niu, Liu, Li, and Huang]{jia2024sceneverse}
Baoxiong Jia, Yixin Chen, Huangyue Yu, Yan Wang, Xuesong Niu, Tengyu Liu, Qing Li, and Siyuan Huang.
\newblock Sceneverse: Scaling 3d vision-language learning for grounded scene understanding.
\newblock In \emph{European Conference on Computer Vision}, pages 289--310. Springer, 2024.

\bibitem[Khirodkar et~al.(2023)Khirodkar, Bansal, Ma, Newcombe, Vo, and Kitani]{khirodkar2023ego}
Rawal Khirodkar, Aayush Bansal, Lingni Ma, Richard Newcombe, Minh Vo, and Kris Kitani.
\newblock Ego-humans: An ego-centric 3d multi-human benchmark.
\newblock In \emph{Proceedings of the IEEE/CVF International Conference on Computer Vision}, pages 19807--19819, 2023.

\bibitem[Kim et~al.(2024)Kim, Pertsch, Karamcheti, Xiao, Balakrishna, Nair, Rafailov, Foster, Lam, Sanketi, et~al.]{kim2024openvla}
Moo~Jin Kim, Karl Pertsch, Siddharth Karamcheti, Ted Xiao, Ashwin Balakrishna, Suraj Nair, Rafael Rafailov, Ethan Foster, Grace Lam, Pannag Sanketi, et~al.
\newblock Openvla: An open-source vision-language-action model.
\newblock \emph{arXiv preprint arXiv:2406.09246}, 2024.

\bibitem[Li et~al.(2024{\natexlab{a}})Li, Ge, Ge, Wang, Wang, Zhang, and Shan]{li2024seed}
Bohao Li, Yuying Ge, Yixiao Ge, Guangzhi Wang, Rui Wang, Ruimao Zhang, and Ying Shan.
\newblock Seed-bench: Benchmarking multimodal large language models.
\newblock In \emph{Proceedings of the IEEE/CVF Conference on Computer Vision and Pattern Recognition}, pages 13299--13308, 2024{\natexlab{a}}.

\bibitem[Li et~al.(2024{\natexlab{b}})Li, Zhang, Guo, Zhang, Li, Zhang, Zhang, Li, Liu, and Li]{li2024llava}
Bo Li, Yuanhan Zhang, Dong Guo, Renrui Zhang, Feng Li, Hao Zhang, Kaichen Zhang, Yanwei Li, Ziwei Liu, and Chunyuan Li.
\newblock Llava-onevision: Easy visual task transfer.
\newblock \emph{arXiv preprint arXiv:2408.03326}, 2024{\natexlab{b}}.

\bibitem[Li et~al.(2024{\natexlab{c}})Li, Zhang, Guo, Zhang, Li, Zhang, Zhang, Li, Liu, and Li]{li2024llavaov}
Bo Li, Yuanhan Zhang, Dong Guo, Renrui Zhang, Feng Li, Hao Zhang, Kaichen Zhang, Yanwei Li, Ziwei Liu, and Chunyuan Li.
\newblock Llava-onevision: Easy visual task transfer.
\newblock \emph{arXiv preprint arXiv:2408.03326}, 2024{\natexlab{c}}.

\bibitem[Li et~al.(2024{\natexlab{d}})Li, Wang, He, Li, Wang, Liu, Wang, Xu, Chen, Luo, et~al.]{li2024mvbench}
Kunchang Li, Yali Wang, Yinan He, Yizhuo Li, Yi Wang, Yi Liu, Zun Wang, Jilan Xu, Guo Chen, Ping Luo, et~al.
\newblock Mvbench: A comprehensive multi-modal video understanding benchmark.
\newblock In \emph{Proceedings of the IEEE/CVF Conference on Computer Vision and Pattern Recognition}, pages 22195--22206, 2024{\natexlab{d}}.

\bibitem[Li et~al.(2025)Li, Zhao, Wang, Wang, Zhou, Srivastava, Gokmen, Lee, Li, Zhang, et~al.]{li2025embodied}
Manling Li, Shiyu Zhao, Qineng Wang, Kangrui Wang, Yu Zhou, Sanjana Srivastava, Cem Gokmen, Tony Lee, Erran~Li Li, Ruohan Zhang, et~al.
\newblock Embodied agent interface: Benchmarking llms for embodied decision making.
\newblock \emph{Advances in Neural Information Processing Systems}, 37:\penalty0 100428--100534, 2025.

\bibitem[Liu et~al.(2024{\natexlab{a}})Liu, Fang, Abbeel, and Levine]{liu2024moka}
Fangchen Liu, Kuan Fang, Pieter Abbeel, and Sergey Levine.
\newblock Moka: Open-vocabulary robotic manipulation through mark-based visual prompting.
\newblock In \emph{First Workshop on Vision-Language Models for Navigation and Manipulation at ICRA 2024}, 2024{\natexlab{a}}.

\bibitem[Liu et~al.(2023)Liu, Li, Wu, and Lee]{liu2023visual}
Haotian Liu, Chunyuan Li, Qingyang Wu, and Yong~Jae Lee.
\newblock Visual instruction tuning.
\newblock In \emph{NeurIPS}, 2023.

\bibitem[Liu et~al.(2024{\natexlab{b}})Liu, Li, Li, and Lee]{liu2023improved}
Haotian Liu, Chunyuan Li, Yuheng Li, and Yong~Jae Lee.
\newblock Improved baselines with visual instruction tuning.
\newblock In \emph{CVPR}, 2024{\natexlab{b}}.

\bibitem[Liu et~al.(2024{\natexlab{c}})Liu, Duan, Zhang, Li, Zhang, Zhao, Yuan, Wang, He, Liu, et~al.]{liu2023mmbench}
Yuan Liu, Haodong Duan, Yuanhan Zhang, Bo Li, Songyang Zhang, Wangbo Zhao, Yike Yuan, Jiaqi Wang, Conghui He, Ziwei Liu, et~al.
\newblock Mmbench: Is your multi-modal model an all-around player?
\newblock In \emph{ECCV}, 2024{\natexlab{c}}.

\bibitem[Lu et~al.(2023)Lu, Bansal, Xia, Liu, Li, Hajishirzi, Cheng, Chang, Galley, and Gao]{lu2023mathvista}
Pan Lu, Hritik Bansal, Tony Xia, Jiacheng Liu, Chunyuan Li, Hannaneh Hajishirzi, Hao Cheng, Kai-Wei Chang, Michel Galley, and Jianfeng Gao.
\newblock Mathvista: Evaluating mathematical reasoning of foundation models in visual contexts.
\newblock In \emph{ICLR}, 2023.

\bibitem[Lu et~al.(2024)Lu, Li, Chen, Xu, Luo, Zhang, and Ye]{lu2024ovis}
Shiyin Lu, Yang Li, Qing-Guo Chen, Zhao Xu, Weihua Luo, Kaifu Zhang, and Han-Jia Ye.
\newblock Ovis: Structural embedding alignment for multimodal large language model.
\newblock \emph{arXiv preprint arXiv:2405.20797}, 2024.

\bibitem[Mangalam et~al.(2023)Mangalam, Akshulakov, and Malik]{mangalam2023egoschema}
Karttikeya Mangalam, Raiymbek Akshulakov, and Jitendra Malik.
\newblock Egoschema: A diagnostic benchmark for very long-form video language understanding.
\newblock \emph{Advances in Neural Information Processing Systems}, 36:\penalty0 46212--46244, 2023.

\bibitem[Masry et~al.(2022)Masry, Long, Tan, Joty, and Hoque]{masry2022chartqa}
Ahmed Masry, Do~Xuan Long, Jia~Qing Tan, Shafiq Joty, and Enamul Hoque.
\newblock Chartqa: A benchmark for question answering about charts with visual and logical reasoning.
\newblock In \emph{ACL}, 2022.

\bibitem[Niu et~al.(2024)Niu, Sharma, Biamby, Quenum, Bai, Shi, Darrell, and Herzig]{niu2024llarva}
Dantong Niu, Yuvan Sharma, Giscard Biamby, Jerome Quenum, Yutong Bai, Baifeng Shi, Trevor Darrell, and Roei Herzig.
\newblock Llarva: Vision-action instruction tuning enhances robot learning.
\newblock \emph{arXiv preprint arXiv:2406.11815}, 2024.

\bibitem[OpenAI(2024)]{OpenAI2024gpt4o}
OpenAI.
\newblock gpt4o, 2024.

\bibitem[Rajbhandari et~al.(2020)Rajbhandari, Rasley, Ruwase, and He]{rajbhandari2020zero}
Samyam Rajbhandari, Jeff Rasley, Olatunji Ruwase, and Yuxiong He.
\newblock Zero: Memory optimizations toward training trillion parameter models.
\newblock In \emph{SC20: International Conference for High Performance Computing, Networking, Storage and Analysis}, pages 1--16. IEEE, 2020.

\bibitem[Rudman et~al.(2025)Rudman, Golovanesky, Bar, Palit, LeCun, Eickhoff, and Singh]{rudman2025forgotten}
William Rudman, Michal Golovanesky, Amir Bar, Vedant Palit, Yann LeCun, Carsten Eickhoff, and Ritambhara Singh.
\newblock Forgotten polygons: Multimodal large language models are shape-blind.
\newblock \emph{arXiv preprint arXiv:2502.15969}, 2025.

\bibitem[Song et~al.(2022)Song, Kil, Pan, Sadler, Chao, and Su]{song2022one}
Chan~Hee Song, Jihyung Kil, Tai-Yu Pan, Brian~M Sadler, Wei-Lun Chao, and Yu Su.
\newblock One step at a time: Long-horizon vision-and-language navigation with milestones.
\newblock In \emph{Proceedings of the IEEE/CVF conference on computer vision and pattern recognition}, pages 15482--15491, 2022.

\bibitem[Suglia et~al.(2021)Suglia, Gao, Thomason, Thattai, and Sukhatme]{suglia2021embodied}
Alessandro Suglia, Qiaozi Gao, Jesse Thomason, Govind Thattai, and Gaurav Sukhatme.
\newblock Embodied bert: A transformer model for embodied, language-guided visual task completion.
\newblock \emph{arXiv preprint arXiv:2108.04927}, 2021.

\bibitem[Team et~al.(2023)Team, Anil, Borgeaud, Alayrac, Yu, Soricut, Schalkwyk, Dai, Hauth, Millican, et~al.]{team2023gemini}
Gemini Team, Rohan Anil, Sebastian Borgeaud, Jean-Baptiste Alayrac, Jiahui Yu, Radu Soricut, Johan Schalkwyk, Andrew~M Dai, Anja Hauth, Katie Millican, et~al.
\newblock Gemini: a family of highly capable multimodal models.
\newblock \emph{arXiv preprint arXiv:2312.11805}, 2023.

\bibitem[Tong et~al.(2024{\natexlab{a}})Tong, Brown, Wu, Woo, Middepogu, Akula, Yang, Yang, Iyer, Pan, et~al.]{tong2024cambrian}
Shengbang Tong, Ellis Brown, Penghao Wu, Sanghyun Woo, Manoj Middepogu, Sai~Charitha Akula, Jihan Yang, Shusheng Yang, Adithya Iyer, Xichen Pan, et~al.
\newblock Cambrian-1: A fully open, vision-centric exploration of multimodal llms.
\newblock In \emph{NeurIPS}, 2024{\natexlab{a}}.

\bibitem[Tong et~al.(2024{\natexlab{b}})Tong, Fan, Zhu, Xiong, Chen, Sinha, Rabbat, LeCun, Xie, and Liu]{tong2024metamorph}
Shengbang Tong, David Fan, Jiachen Zhu, Yunyang Xiong, Xinlei Chen, Koustuv Sinha, Michael Rabbat, Yann LeCun, Saining Xie, and Zhuang Liu.
\newblock Metamorph: Multimodal understanding and generation via instruction tuning.
\newblock \emph{arXiv preprint arXiv:2412.14164}, 2024{\natexlab{b}}.

\bibitem[Tong et~al.(2024{\natexlab{c}})Tong, Liu, Zhai, Ma, LeCun, and Xie]{tong2024eyes}
Shengbang Tong, Zhuang Liu, Yuexiang Zhai, Yi Ma, Yann LeCun, and Saining Xie.
\newblock Eyes wide shut? exploring the visual shortcomings of multimodal llms.
\newblock In \emph{CVPR}, 2024{\natexlab{c}}.

\bibitem[Wang et~al.(2024)Wang, Luo, Chen, Mai, Guo, Dong, Li, Ma, Gao, et~al.]{wang2024mllm}
Chenyu Wang, Weixin Luo, Qianyu Chen, Haonan Mai, Jindi Guo, Sixun Dong, Zhengxin Li, Lin Ma, Shenghua Gao, et~al.
\newblock Mllm-tool: A multimodal large language model for tool agent learning.
\newblock \emph{arXiv preprint arXiv:2401.10727}, 2024.

\bibitem[Wang et~al.(2023)Wang, Wei, Schuurmans, Le, Chi, Narang, Chowdhery, and Zhou]{wang2022self}
Xuezhi Wang, Jason Wei, Dale Schuurmans, Quoc~V Le, Ed~H. Chi, Sharan Narang, Aakanksha Chowdhery, and Denny Zhou.
\newblock Self-consistency improves chain of thought reasoning in language models.
\newblock In \emph{ICLR}, 2023.

\bibitem[Wei et~al.(2022)Wei, Wang, Schuurmans, Bosma, Xia, Chi, Le, Zhou, et~al.]{wei2022chain}
Jason Wei, Xuezhi Wang, Dale Schuurmans, Maarten Bosma, Fei Xia, Ed Chi, Quoc~V Le, Denny Zhou, et~al.
\newblock Chain-of-thought prompting elicits reasoning in large language models.
\newblock In \emph{NeurIPS}, 2022.

\bibitem[Wu et~al.(2024)Wu, Chen, Pan, Liu, Liu, Dai, Gao, Ma, Wu, Wang, et~al.]{wu2024deepseek}
Zhiyu Wu, Xiaokang Chen, Zizheng Pan, Xingchao Liu, Wen Liu, Damai Dai, Huazuo Gao, Yiyang Ma, Chengyue Wu, Bingxuan Wang, et~al.
\newblock Deepseek-vl2: Mixture-of-experts vision-language models for advanced multimodal understanding.
\newblock \emph{arXiv preprint arXiv:2412.10302}, 2024.

\bibitem[Xu et~al.(2024)Xu, Zhao, Wang, Liu, Ma, and Gao]{xu2024cad}
Jingwei Xu, Zibo Zhao, Chenyu Wang, Wen Liu, Yi Ma, and Shenghua Gao.
\newblock Cad-mllm: Unifying multimodality-conditioned cad generation with mllm.
\newblock \emph{arXiv preprint arXiv:2411.04954}, 2024.

\bibitem[Yang et~al.(2024{\natexlab{a}})Yang, Yang, Zhang, Hui, Zheng, Yu, Li, Liu, Huang, Wei, et~al.]{yang2024qwen2}
An Yang, Baosong Yang, Beichen Zhang, Binyuan Hui, Bo Zheng, Bowen Yu, Chengyuan Li, Dayiheng Liu, Fei Huang, Haoran Wei, et~al.
\newblock Qwen2. 5 technical report.
\newblock \emph{arXiv preprint arXiv:2412.15115}, 2024{\natexlab{a}}.

\bibitem[Yang et~al.(2024{\natexlab{b}})Yang, Yang, Gupta, Han, Fei-Fei, and Xie]{yang2024thinking}
Jihan Yang, Shusheng Yang, Anjali~W Gupta, Rilyn Han, Li Fei-Fei, and Saining Xie.
\newblock Thinking in space: How multimodal large language models see, remember, and recall spaces.
\newblock \emph{arXiv preprint arXiv:2412.14171}, 2024{\natexlab{b}}.

\bibitem[Yang et~al.(2024{\natexlab{c}})Yang, Zhang, Shao, Zhang, Bin, Wang, and Luo]{yang2024dynamic}
Yue Yang, Shuibai Zhang, Wenqi Shao, Kaipeng Zhang, Yi Bin, Yu Wang, and Ping Luo.
\newblock Dynamic multimodal evaluation with flexible complexity by vision-language bootstrapping.
\newblock \emph{arXiv preprint arXiv:2410.08695}, 2024{\natexlab{c}}.

\bibitem[Yu et~al.(2025)Yu, Li, Wang, Chen, and Zhu]{yu2025inst3dlmminstanceaware3dscene}
Hanxun Yu, Wentong Li, Song Wang, Junbo Chen, and Jianke Zhu.
\newblock Inst3d-lmm: Instance-aware 3d scene understanding with multi-modal instruction tuning, 2025.

\bibitem[Yu et~al.(2019)Yu, Chen, Gkioxari, Bansal, Berg, and Batra]{yu2019multi}
Licheng Yu, Xinlei Chen, Georgia Gkioxari, Mohit Bansal, Tamara~L Berg, and Dhruv Batra.
\newblock Multi-target embodied question answering.
\newblock In \emph{Proceedings of the IEEE/CVF Conference on Computer Vision and Pattern Recognition}, pages 6309--6318, 2019.

\bibitem[Yue et~al.(2024{\natexlab{a}})Yue, Ni, Zhang, Zheng, Liu, Zhang, Stevens, Jiang, Ren, Sun, et~al.]{yue2023mmmu}
Xiang Yue, Yuansheng Ni, Kai Zhang, Tianyu Zheng, Ruoqi Liu, Ge Zhang, Samuel Stevens, Dongfu Jiang, Weiming Ren, Yuxuan Sun, et~al.
\newblock Mmmu: A massive multi-discipline multimodal understanding and reasoning benchmark for expert agi.
\newblock In \emph{CVPR}, 2024{\natexlab{a}}.

\bibitem[Yue et~al.(2024{\natexlab{b}})Yue, Ni, Zhang, Zheng, Liu, Zhang, Stevens, Jiang, Ren, Sun, et~al.]{yue2024mmmu}
Xiang Yue, Yuansheng Ni, Kai Zhang, Tianyu Zheng, Ruoqi Liu, Ge Zhang, Samuel Stevens, Dongfu Jiang, Weiming Ren, Yuxuan Sun, et~al.
\newblock Mmmu: A massive multi-discipline multimodal understanding and reasoning benchmark for expert agi.
\newblock In \emph{Proceedings of the IEEE/CVF Conference on Computer Vision and Pattern Recognition}, pages 9556--9567, 2024{\natexlab{b}}.

\bibitem[Zhai et~al.(2024)Zhai, Bai, Lin, Pan, Tong, Zhou, Suhr, Xie, LeCun, Ma, et~al.]{zhai2024fine}
Yuexiang Zhai, Hao Bai, Zipeng Lin, Jiayi Pan, Shengbang Tong, Yifei Zhou, Alane Suhr, Saining Xie, Yann LeCun, Yi Ma, et~al.
\newblock Fine-tuning large vision-language models as decision-making agents via reinforcement learning.
\newblock In \emph{NeurIPS}, 2024.

\bibitem[Zhang et~al.(2025)Zhang, Yao, Pi, Liang, et~al.]{zhang2025vlm}
Jianshu Zhang, Dongyu Yao, Renjie Pi, Paul~Pu Liang, et~al.
\newblock Vlm 2-bench: A closer look at how well vlms implicitly link explicit matching visual cues.
\newblock \emph{arXiv preprint arXiv:2502.12084}, 2025.

\bibitem[Zhang et~al.(2024)Zhang, Li, Liu, Lee, Gui, Fu, Feng, Liu, and Li]{zhang2024llavanextvideo}
Yuanhan Zhang, Bo Li, haotian Liu, Yong~jae Lee, Liangke Gui, Di Fu, Jiashi Feng, Ziwei Liu, and Chunyuan Li.
\newblock Llava-next: A strong zero-shot video understanding model, 2024.

\bibitem[Zhou et~al.(2024)Zhou, Yu, Babu, Tirumala, Yasunaga, Shamis, Kahn, Ma, Zettlemoyer, and Levy]{zhou2024transfusion}
Chunting Zhou, Lili Yu, Arun Babu, Kushal Tirumala, Michihiro Yasunaga, Leonid Shamis, Jacob Kahn, Xuezhe Ma, Luke Zettlemoyer, and Omer Levy.
\newblock Transfusion: Predict the next token and diffuse images with one multi-modal model.
\newblock \emph{arXiv preprint arXiv:2408.11039}, 2024.

\bibitem[Zhu et~al.(2024{\natexlab{a}})Zhu, Wang, Zhang, Pang, and Liu]{zhu2024llava}
Chenming Zhu, Tai Wang, Wenwei Zhang, Jiangmiao Pang, and Xihui Liu.
\newblock Llava-3d: A simple yet effective pathway to empowering lmms with 3d-awareness.
\newblock \emph{arXiv preprint arXiv:2409.18125}, 2024{\natexlab{a}}.

\bibitem[Zhu et~al.(2024{\natexlab{b}})Zhu, Wang, Zhao, Xu, and Xie]{zhu2024dynamic}
Kaijie Zhu, Jindong Wang, Qinlin Zhao, Ruochen Xu, and Xing Xie.
\newblock Dynamic evaluation of large language models by meta probing agents.
\newblock \emph{arXiv preprint arXiv:2402.14865}, 2024{\natexlab{b}}.

\end{thebibliography}
}
\clearpage
\setcounter{page}{1}
\maketitlesupplementary

In these supplementary materials, we provide the following:
\begin{itemize}
\item Details on the construction and annotation pipeline of \textit{All-Angles Bench} (Section \ref{sec:sup_construction_pipe}); 
\item Evaluation setup, implementation detail of CoT methods, evaluation results visualization and complete evaluation results for the tiny \textit{All-Angles Bench}  (Section \ref{sec:sup_exp});
\item Additional visualization results and prompts (Section \ref{sec:sup_vis}).
\end{itemize}

\section{Construction and Annotation Pipeline}
\label{sec:sup_construction_pipe}

\subsection{Dataset Collection}
We manually selected 83 scenes from Ego-Exo4D~\cite{grauman2024ego} and 7 scenes from EgoHumans~\cite{khirodkar2023ego} to ensure the diversity of scenes. Given the high density of viewpoints in some EgoHumans scenes, we carefully curated a subset of more spatially dispersed views to avoid excessive redundancy. As a result, we retained 4 - 5 views per scene. All multi-view images were standardized to a resolution of 796 × 448 pixels.

\subsection{Question Creation}
For each generated question, we recorded the following key attributes: question index, source dataset, task category, image list path, question text, and multiple-choice options. Since the questions would undergo a human-in-the-loop quality review and verification process, letter-based answer choices were not generated at this stage. 

Among the six task categories in \textit{All-Angles Bench}, five were generated using an MLLM~\cite{OpenAI2024gpt4o}. For the \textbf{Camera Pose Estimation} task, however, we designed a dedicated question template to structure the question generation process. The system prompt, task-specific prompts for the five generated tasks, and the camera pose estimation question template are illustrated in Figures \ref{fig:system_prompts}, \ref{fig:task_specific_prompts}, and \ref{fig:cam_pose_template}, respectively.

\begin{figure*}
\begin{tcolorbox}
We first have to say thank for your efforts in annotating the \textit{All-Angles Bench} benchmark. Your contribution is invaluable and plays a crucial role in advancing this research. \\\\
\textbf{Your Task:}\\
Review, refine, and finalize \textbf{multiple-choice questions (MCQs)} generated by an MLLM. Ensure they are \textbf{clear, accurate, and correctly answered}.\\

\textbf{What to Check:}

\begin{enumerate}
\item \textit{Clarity \& Relevance} – The question must match the image/context and be easy to understand.
\item \textit{Grammar \& Precision} – No errors or vague phrasing.
\item \textit{Answer Quality} – The correct answer must be \textbf{factually accurate}, and distractors should be \textbf{plausible yet clearly incorrect}.\\
\end{enumerate}

\textbf{Example Before \& After:}

\textit{Original Question:}
\begin{quote}
``What if the man with black pants in camera View 1 starts walking toward the view, what would be the trajectory of the man in camera View 3?"
\end{quote}

\noindent \textcolor{red}{\textbf{\ding{55}}} \textit{Issue:} ``toward the view" is unclear.

\textit{Revised Question:}
\begin{quote}
``What if the man with black pants in camera View 1 starts walking toward the camera position of the view, what would be the trajectory of the man in camera View 3?"
\end{quote}

\noindent \textcolor{green}{\checkmark} \textit{Fix:} Provide a more specific description of the subject's movement.\\
\noindent \textcolor{green}{\checkmark} \textit{Final Answer:} (B) walking toward the right side of View 3.\\

\textbf{How to Annotate}
\begin{enumerate}
    \item \textbf{Verify} the questions and options for plausibility and accuracy based on the images/data provided.
    \item \textbf{Correct} unclear wording, grammatical errors, or inaccuracies in the questions.
    \item \textbf{Finalize} the correct answer.\\
\end{enumerate}

\textbf{Common Mistakes to Avoid}\\
\noindent \textcolor{red}{\textbf{\ding{55}}} Vague phrasing (e.g., ``toward the view").\\
\noindent \textcolor{red}{\textbf{\ding{55}}} Misleading or contradictory options.\\
\noindent \textcolor{red}{\textbf{\ding{55}}} MLLM hallucinations (made-up facts or objects)

\end{tcolorbox}
\caption{\textbf{The streamlined version of annotation guideline for annotators to follow.} It outlines key verification steps, common pitfalls, and examples to help annotators improve question clarity, accuracy, and answer quality.}
\label{fig:annotation_guideline}
\vspace{-15pt}
\end{figure*}

\begin{figure*}[!h]
    \centering
    \includegraphics[width=\linewidth]{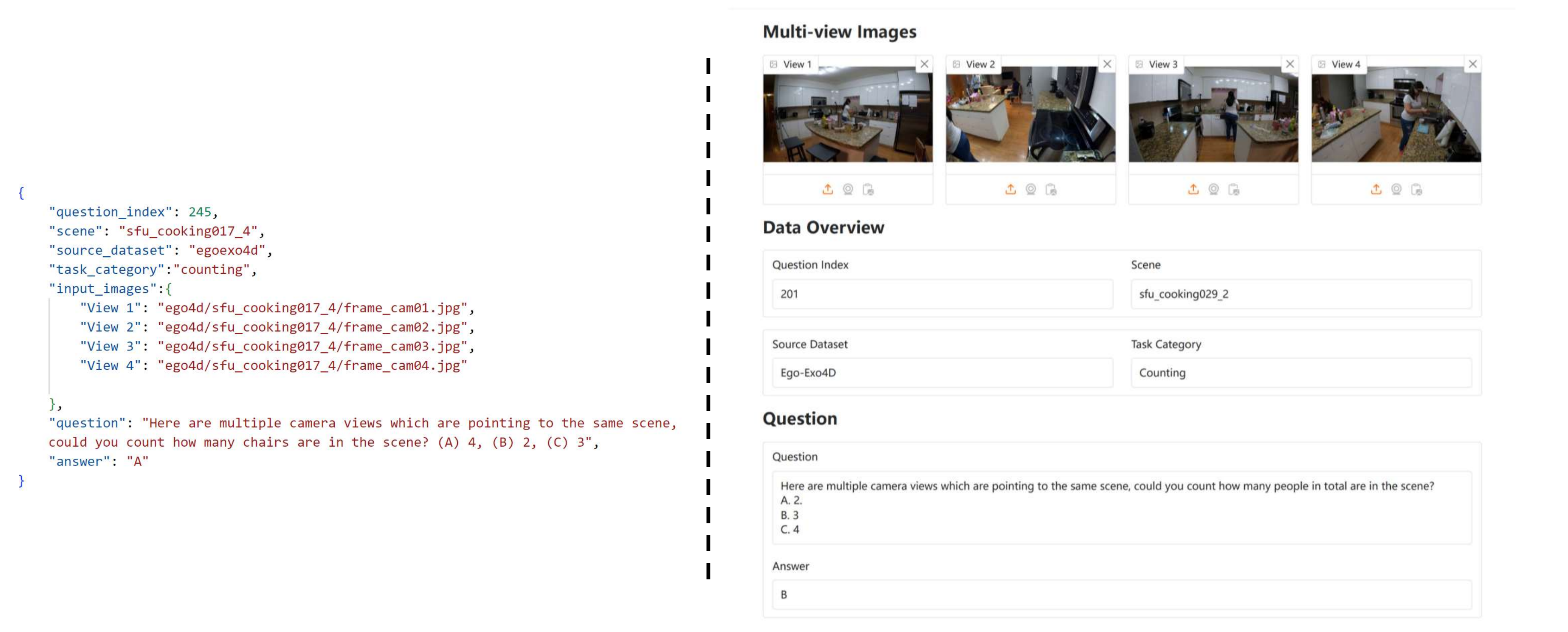}
    \caption{\textbf{Left:} A structured JSON representation of a question-answer pair. \textbf{Right:} A snapshot of the GUI-based Annotation Platform used for reviewing and refining annotations. Best viewed zoomed in for details.}
    \label{fig:bench_json_ui}
\end{figure*}

\begin{figure}[!h]
    \centering
    \includegraphics[width=\linewidth]{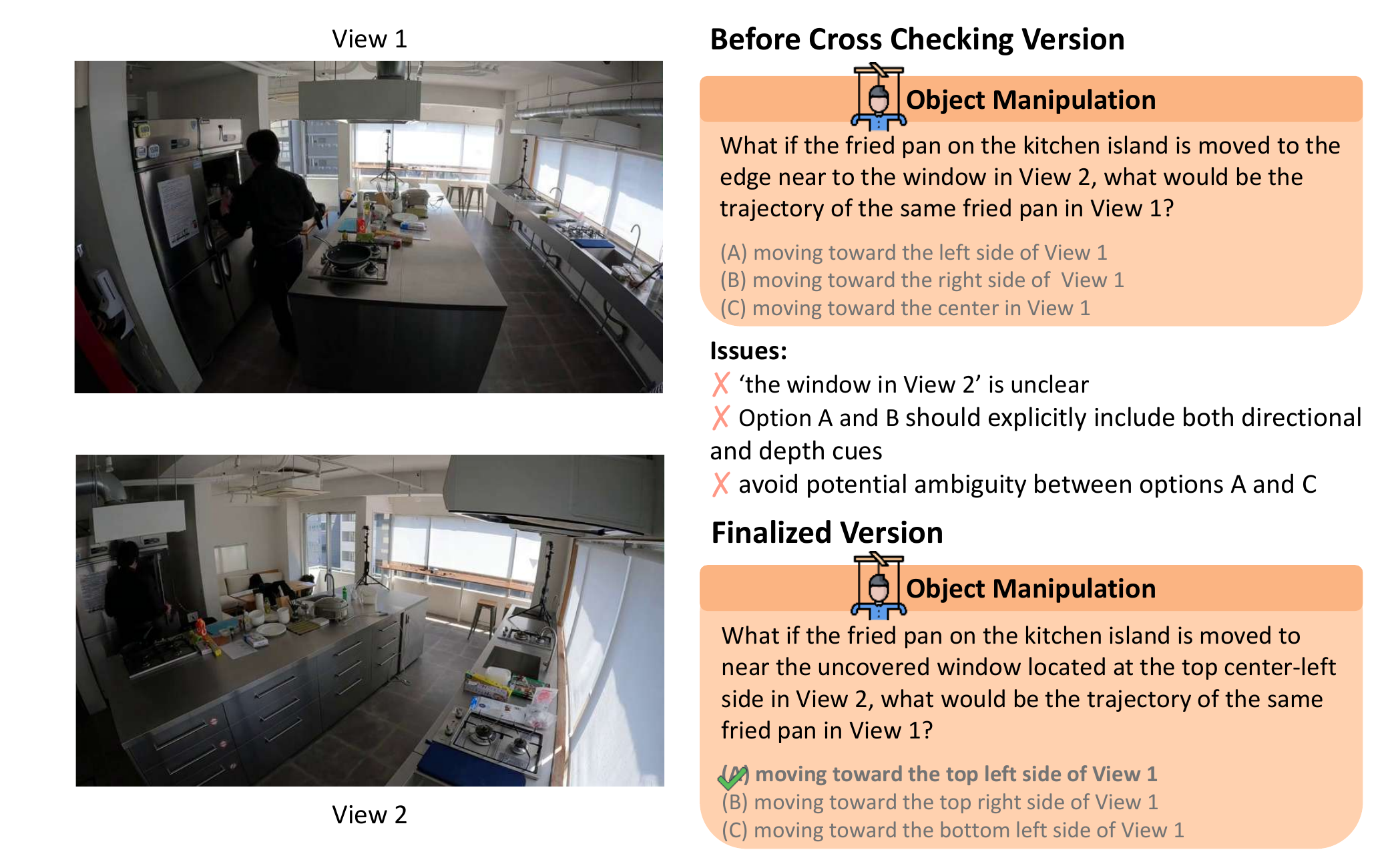}
    \caption{\textbf{Comparison of a flagged ambiguous case before modification and its finalized version after cross-checking.} The initial annotation was reviewed by multiple annotators, with ambiguities resolved through discussions to ensure clarity and consistency.}
    \label{fig:cross_check}
\end{figure}

\subsection{Human Annotation and Quality Check}
In this study, we hired \textbf{eight Ph.D. students in STEM fields} specializing in natural language processing or computer vision, to manually annotate all questions. To ensure consistency, we provided a comprehensive annotation guideline, which was refined and structured into a streamlined version, as illustrated in Figure \ref{fig:annotation_guideline}. 

Additionally, to maintain high annotation quality, we implemented a multi-stage verification process. Before the formal annotation process, annotators were required to complete a set of sample questions to familiarize themselves with the standards and guidelines. To minimize errors and ambiguities, each annotation was then cross-checked by at least one other annotator, with any disagreements resolved through group discussions. Figure \ref{fig:cross_check} presents an example comparing an initial annotation with its final version after cross-checking. Such ambiguous or unclear instances were flagged for review and collaboratively examined in team meetings, ensuring a standardized and consistent annotation process. Furthermore, we adopted a random sampling review mechanism, periodically evaluating a subset of annotated data to ensure strict adherence to the guidelines. The finalized benchmark is stored in \texttt{JSON} format, with an example visualization provided in Figure \ref{fig:bench_json_ui}.

To streamline the annotation and quality control process, we developed a GUI-based annotation platform, as shown in Figure \ref{fig:bench_json_ui}. This platform provides an intuitive interface enabling annotators to inspect and edit annotations efficiently. Annotators can seamlessly browse multi-view input images, modify questions, and adjust answer options with ease, ensuring both accuracy and consistency in the annotations.

Notably, the entire dataset collection and processing required \textbf{over 300 person-hours}, reflecting our meticulous attention to detail in ensuring the benchmark’s high reliability and quality for the relevant research community.

\section{Experiment Details}
\label{sec:sup_exp}

\subsection{Evaluation Setup}
\label{sec:evaluation_setup}
Our evaluation is conducted using the \texttt{VLMEvalKit}~\cite{duan2024vlmevalkit} framework. In order to ensure the reproducibility of our evaluation, we employ a greedy decoding strategy for all models, setting the temperature to 0 unless otherwise stated. The text input follows a standardized format: \texttt{[Question][Options][Post-prompt]}, where the post-prompt instructs: ``\textit{Answer with the option's letter from the given choices directly.}" 

To ensure that all final predictions are formatted as single-letter outputs, facilitating subsequent evaluation against the ground-truth answers and minimizing errors due to fuzzy matching, we leverage the open-source LLM, Qwen2.5-32B~\cite{yang2024qwen2} to extract the predicted options accurately. The corresponding prompt is shown in Figure \ref{fig:fetching_prompts}.

For human-level performance evaluation on the tiny 250-question benchmark, we invited two additional Ph.D. students in STEM fields who were not involved in the annotation process to answer the questions. Each evaluator was assigned 125 questions and given unlimited time to answer with their best effort. Their combined scores serve as the human performance baseline for this tiny benchmark.

To eliminate potential biases introduced by contextual cues, we exclude paired data from this subset, preventing evaluators from leveraging strong prior knowledge. Additionally, for questions involving only two views, we ensure consistency with the MLLM setup by displaying only the relevant input views rather than all available ones. To further prevent evaluators from unintentionally deriving answers from sequentially presented images, we randomly shuffle the question order, ensuring independent assessment of each query.

\subsection{Implementation Details of CoT Methods}
Inspired by~\cite{yang2024thinking}, we evaluate three distinct reasoning-based prompting strategies on our benchmark: \textbf{Zero-Shot CoT}, \textbf{Self-Consistency}, and our proposed \textbf{Identification CoT}. Below, we outline the implementation details.

Notably, after generating intermediate reasoning steps and predictions using the three CoT approaches, we apply a standardized post-processing step. Specifically, we leverage an additional open-source LLM to explicitly extract the final answer from the generated response, as described in Section \ref{sec:evaluation_setup}. 

\begin{figure}
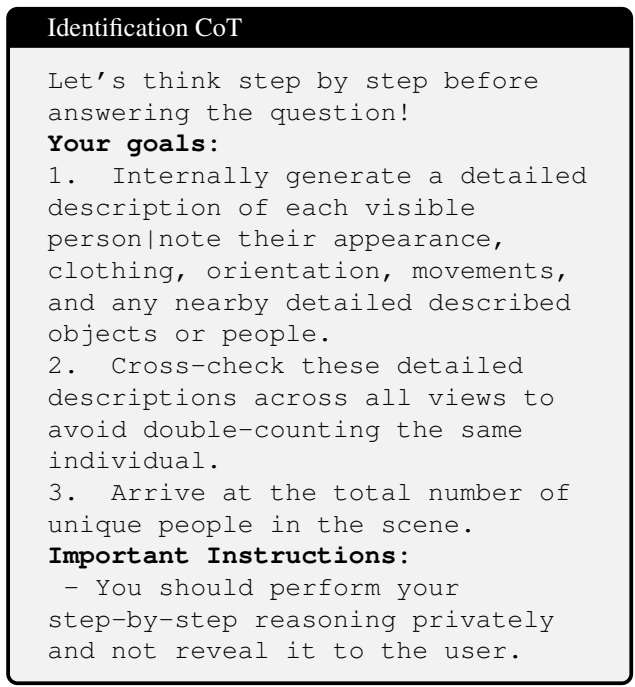

\begin{tcolorbox}[colback=black!5!white,colframe=black!75!black,title=Identification CoT]
\texttt{Let’s think step by step before answering the question!}\\
\texttt{\textbf{Your goals:}}

\texttt{1. Internally generate a detailed description of each visible person—note their appearance, clothing, orientation, movements, and any nearby detailed described objects or people.}

\texttt{2. Cross-check these detailed descriptions across all views to avoid double-counting the same individual.}

\texttt{3. Arrive at the total number of unique people in the scene.}

\texttt{\textbf{Important Instructions:}}

\texttt{ - You should perform your step-by-step reasoning privately and not reveal it to the user.}

\end{tcolorbox}
\caption{\textbf{Our proposed Identification CoT prompt.} To design for counting tasks with partial-visibility, our prompt guides the MLLM to systematically list each target entity across all views.}
\label{fig:id_cot}
\end{figure}

\begin{itemize}
    \item \textbf{Zero-Shot CoT:} Building on prior works~\cite{rajbhandari2020zero,wei2022chain}, we enhance step-by-step reasoning in the MLLM by appending the phrase, ``\textit{Let’s think step by step}", to each question in the post-prompt. The decoding parameter, \textit{temperature}, is set to 0 to ensure deterministic inference.
    \item \textbf{Self-Consistency:} Following the Self Consistency approach~\cite{wang2022self}, we prompt MLLMs to generate multiple independent responses for each question. To encourage diversity, we set \textit{temperature} to 0.6 and conduct five independent inference runs, selecting the most frequently occurring prediction as the final answer.
    \item \textbf{Identification CoT:} Designed specifically for counting tasks, Identification CoT prompts the MLLM to list each target entity visible across all views, mitigating the risk of double-counting and improving accuracy. The corresponding prompt is shown in Figure \ref{fig:id_cot}. This method adopts the same settings as Zero-Shot CoT, using \textit{temperature} equals 0 and a single inference pass to generate the final prediction.
    
\end{itemize}

\begin{figure*}[!h]
    \centering
    \includegraphics[width=\linewidth]{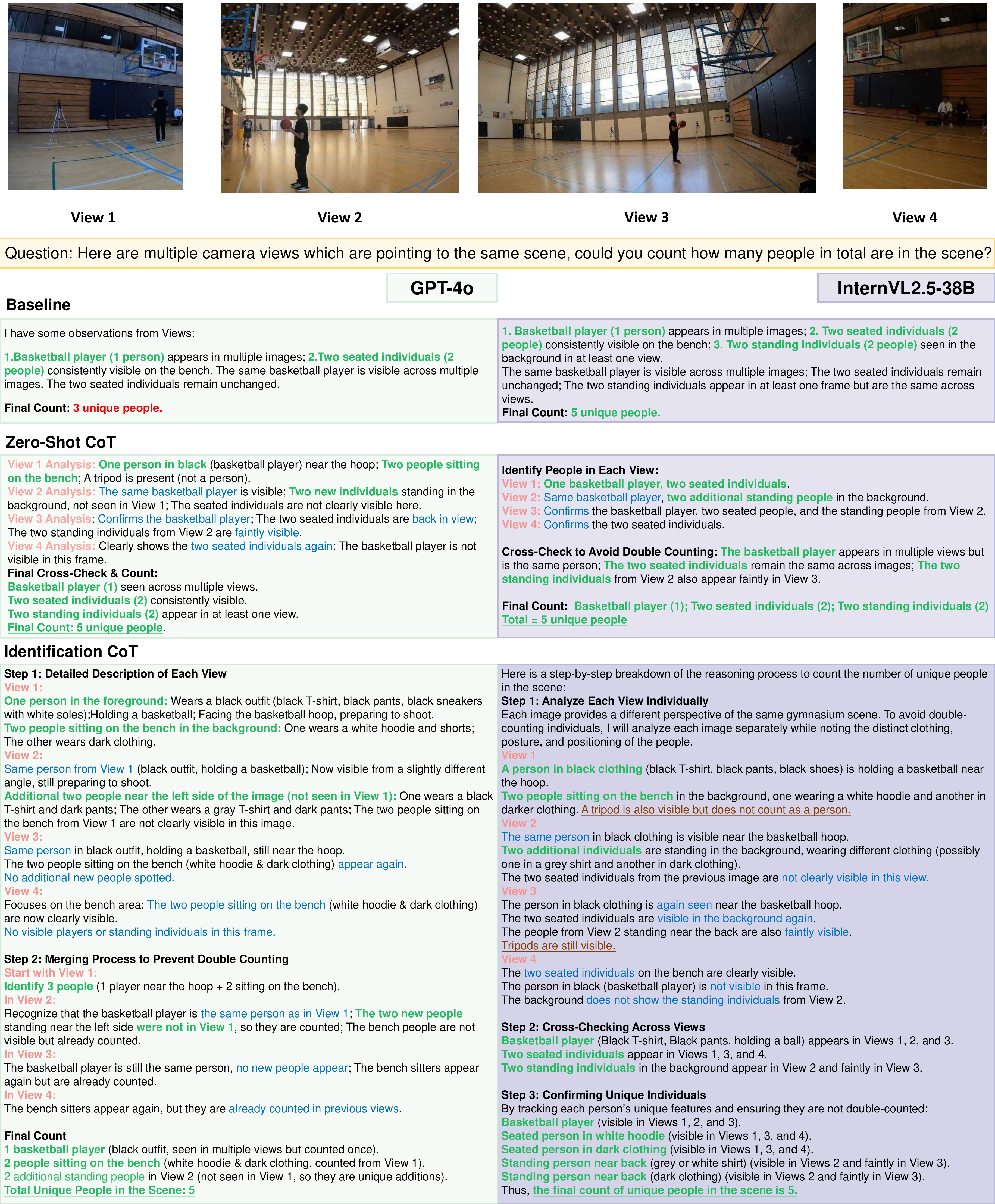}
    \caption{Comparison of model outputs on the same questions under different prompting methods. (Case 1)}
    \label{fig:cot_case1}
\end{figure*}

\begin{figure*}[!h]
    \centering
    \includegraphics[width=\linewidth]{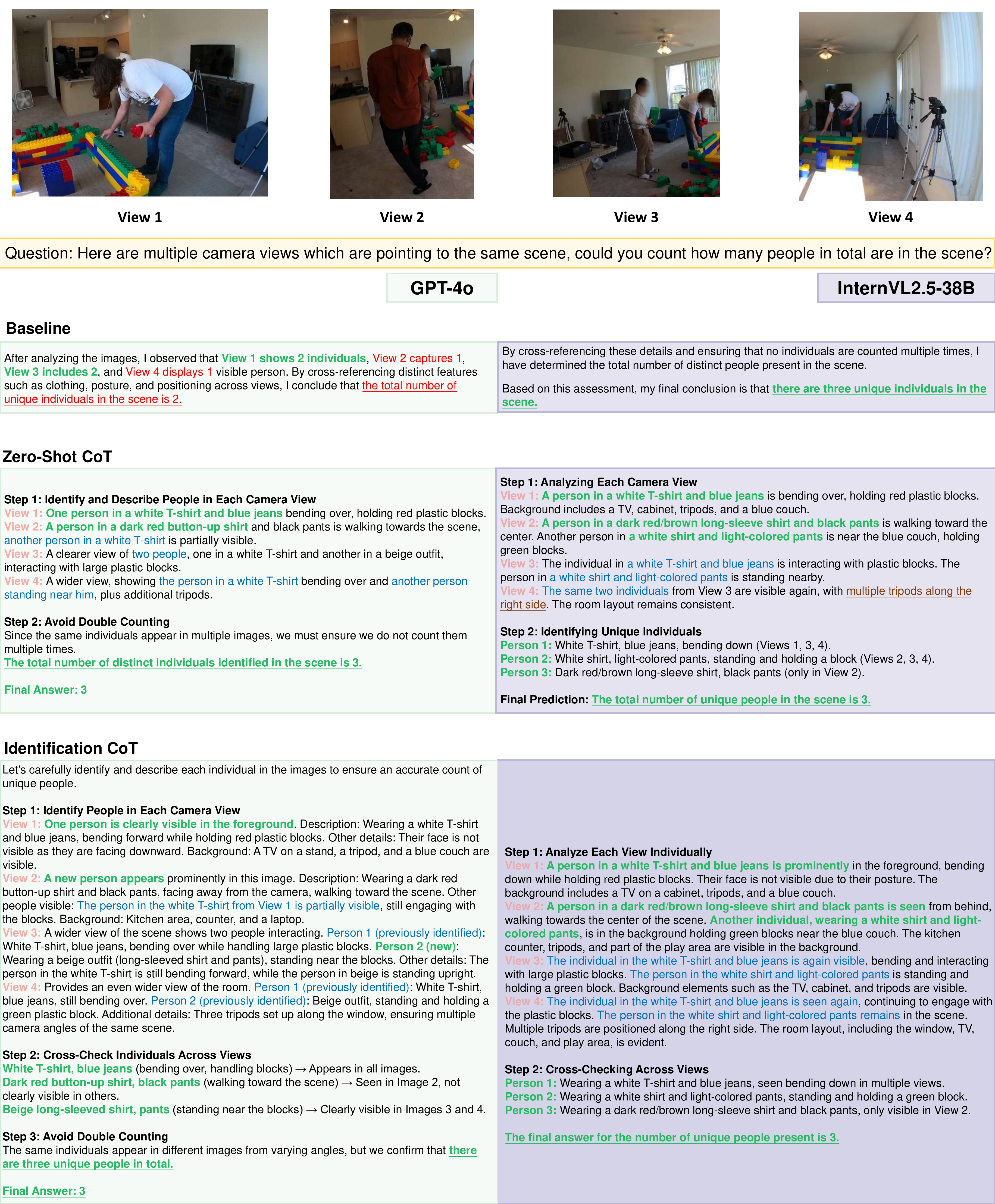}
    \caption{Comparison of model outputs on the same questions under different prompting methods. (Case 2)}
    \label{fig:cot_case2}
\end{figure*}

Figures \ref{fig:cot_case1} and \ref{fig:cot_case2} illustrate model outputs for two MLLMs --- GPT-4o~\cite{OpenAI2024gpt4o} and InternVL2.5-38B~\cite{chen2024expanding} --- under the three prompting strategies.

\subsection{Evaluation Results Visualization}
Figure \ref{fig:confusion} visualizes model performance across six task categories, where color intensity represents precision levels ---cooler colors indicate lower accuracy, while warmer colors denote higher accuracy.

\subsection{More Evaluation Results}
Table \ref{tab:tiny_dataset_table} presents the evaluation results of 27 MLLMs, encompassing both closed-source and open-source models on the 250-question benchmark. The findings remain consistent with those in the main text, confirming that human performance significantly surpasses that of all MLLMs.

\section{Visualization Results}
\label{sec:sup_vis}
In this section, we present a comprehensive visualization of the scene across all available views for convenience and consistency. While some questions do not require every view, we ensure that only the relevant ones are provided as input during inference.

\subsection{Benchmark Examples}
\begin{figure*}[!h]
    \centering
    \includegraphics[width=\linewidth]{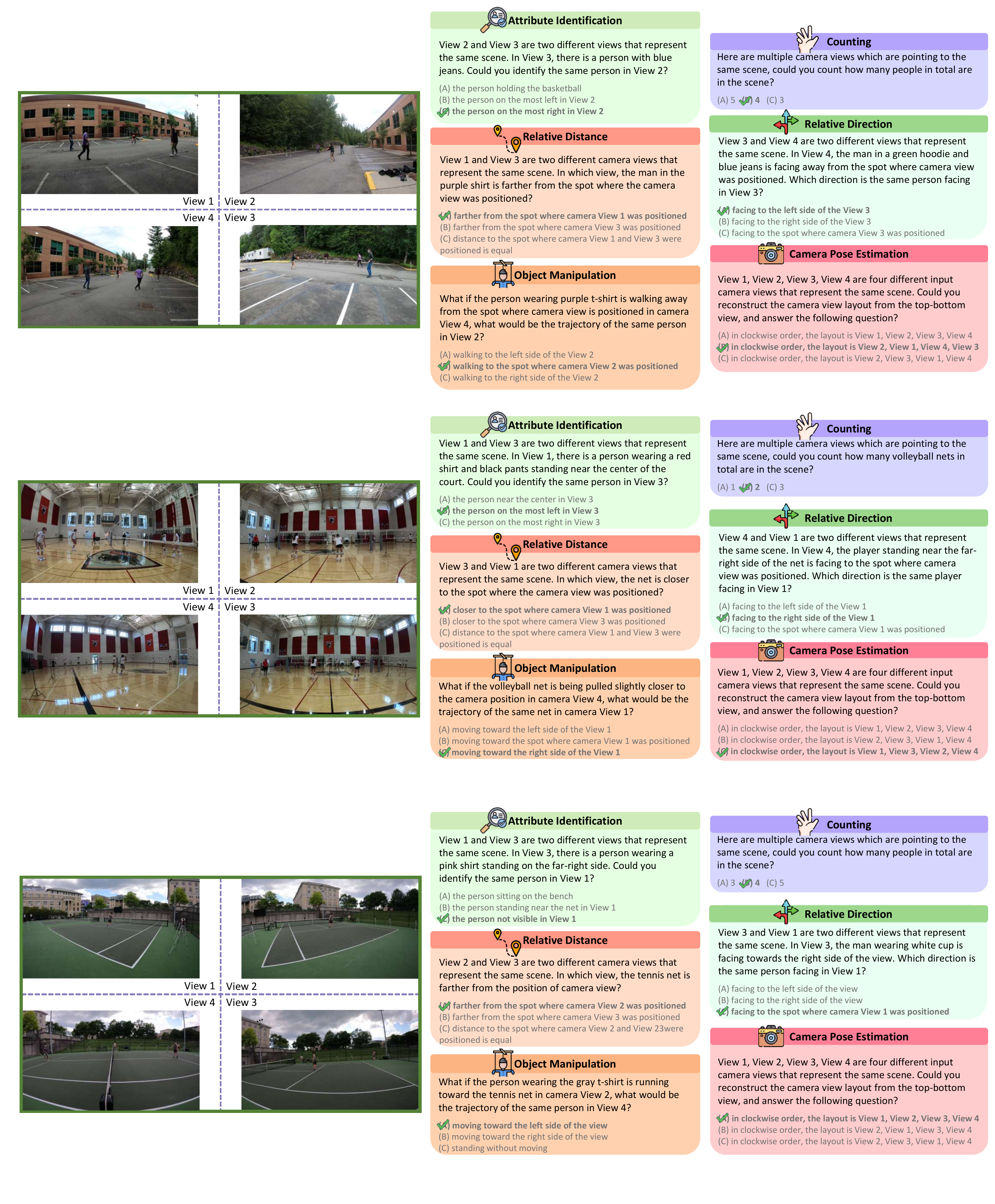}
    \caption{\textit{All-Angles Bench} Samples (Part I)}
    \label{fig:bench_vis_part1}
\end{figure*}

\begin{figure*}[!h]
    \centering
    \includegraphics[width=\linewidth]{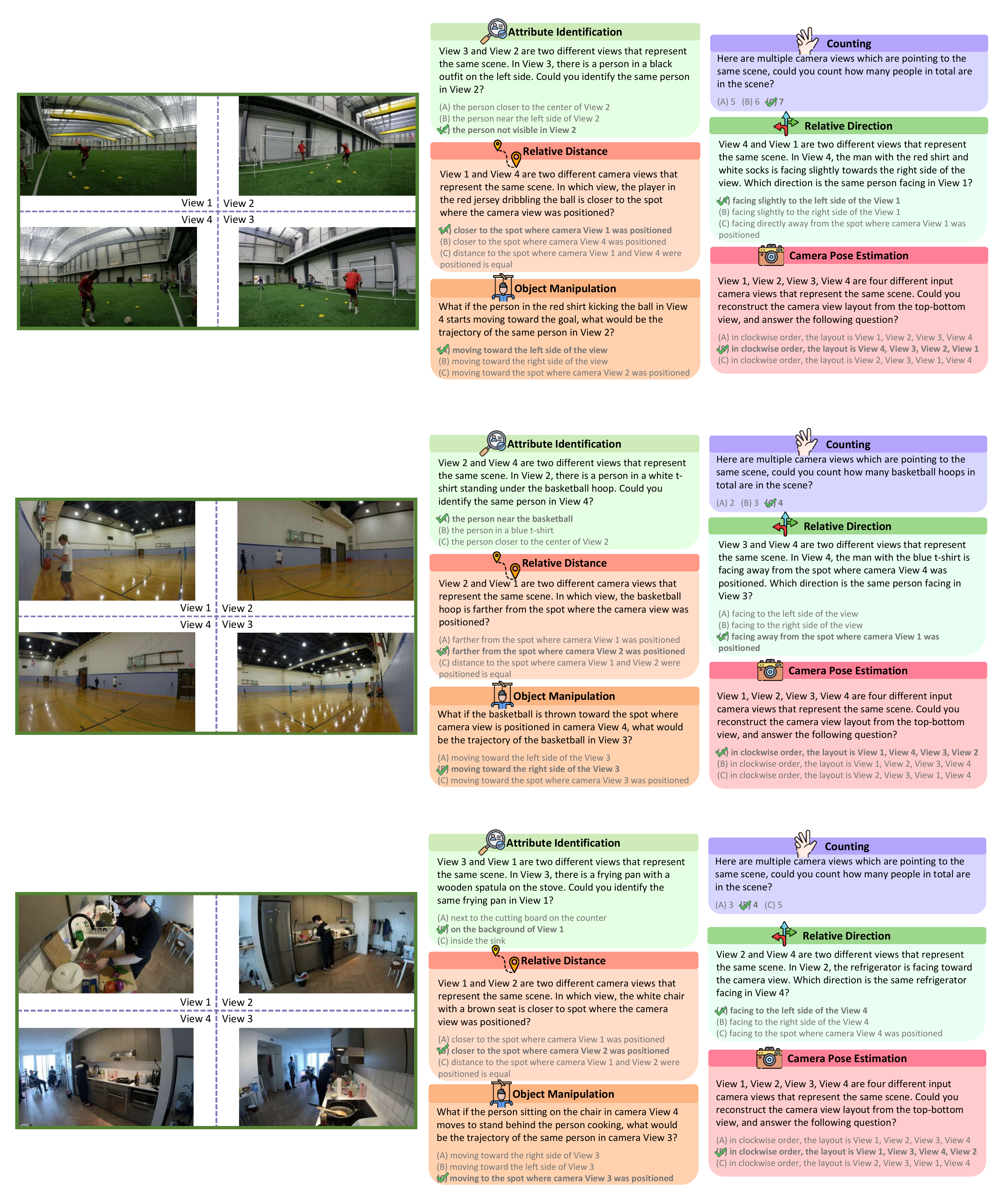}
    \caption{\textit{All-Angles Bench} Samples (Part II)}
    \label{fig:bench_vis_part2}
\end{figure*}

Figures \ref{fig:bench_vis_part1} and \ref{fig:bench_vis_part2} showcase additional primary question-and-answer examples. These illustrations highlight the multi-view image inputs alongside their corresponding tasks and Q\&A pairs, demonstrating the diversity and complexity of our benchmark.

\subsection{Pair Data Examples}
\begin{figure*}[!h]
    \centering
    \includegraphics[width=\linewidth]{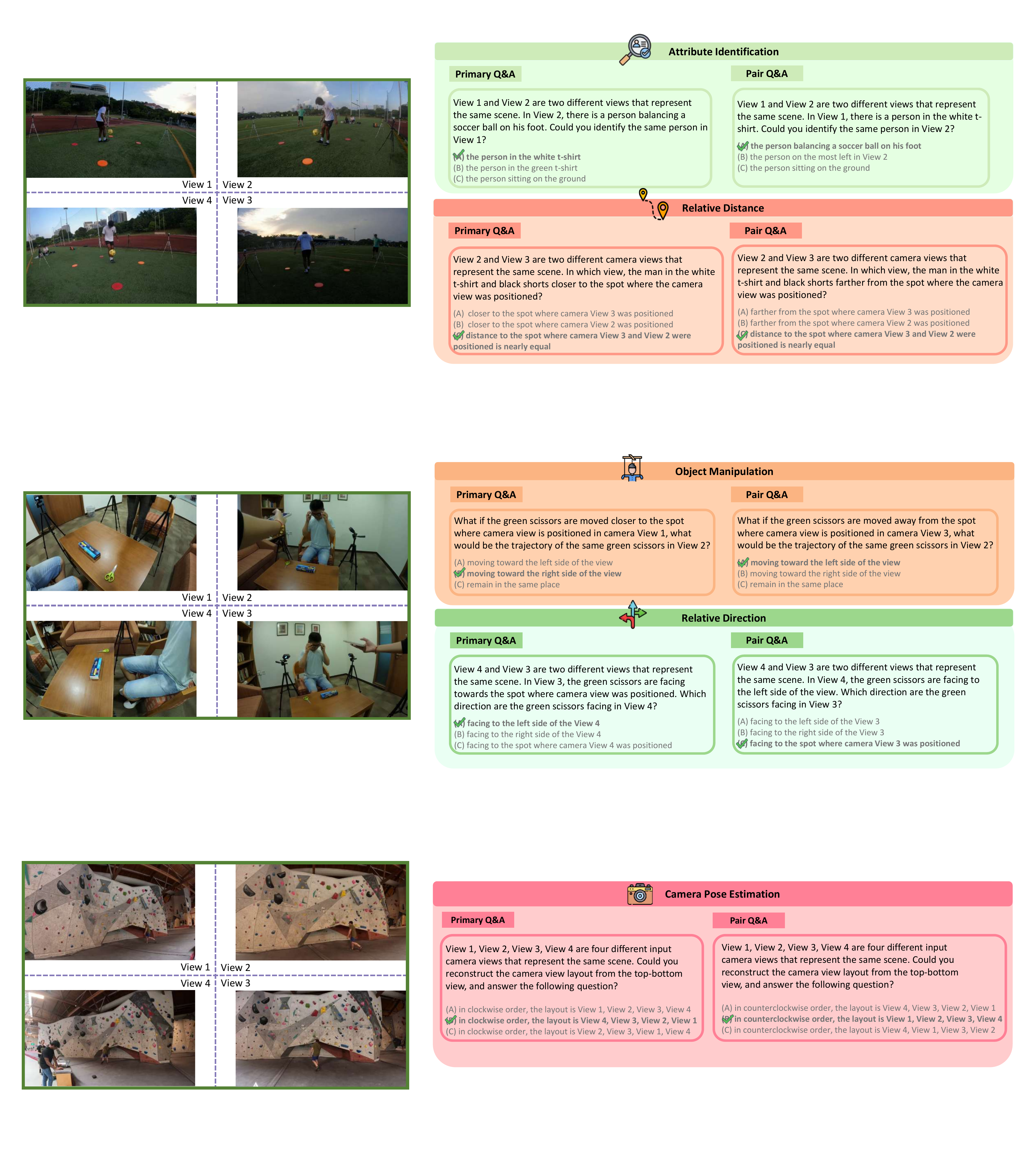}
    \caption{Paired Data Samples (Part I)}
    \label{fig:pair_data_part1}
\end{figure*}

\begin{figure*}[!h]
    \centering
    \includegraphics[width=\linewidth]{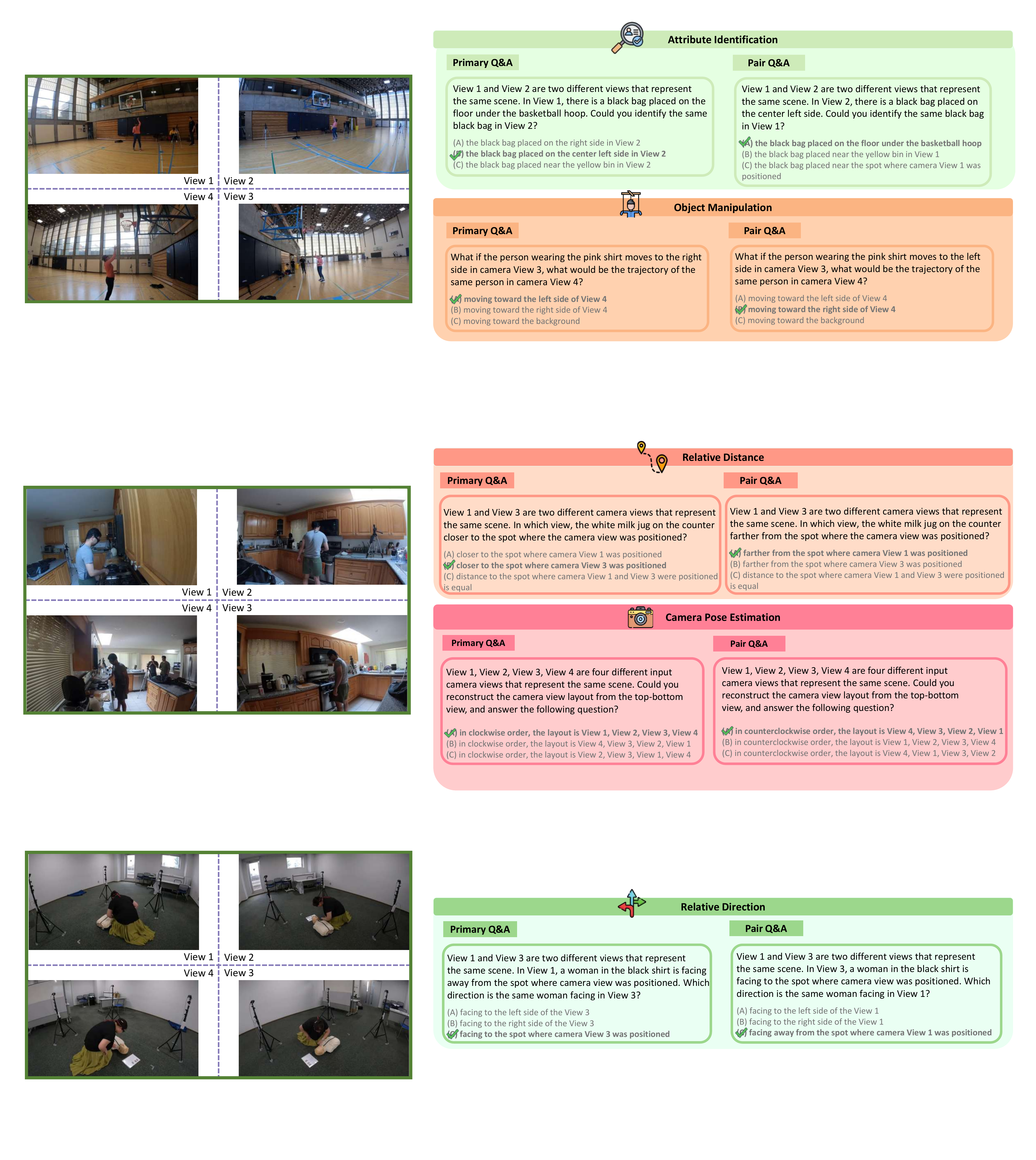}
    \caption{Paired Data Samples (Part II)}
    \label{fig:pair_data_part2}
\end{figure*}

Figures \ref{fig:pair_data_part1} and \ref{fig:pair_data_part2} provide more examples of generated paired data. Each figure presents both the \textbf{primary Q\&A} and the generated \textbf{pair Q\&A}, presented side by side to illustrate their structural alignment and transformation process. This comparison emphasizes how the paired Q\&A is derived from the primary one, reinforcing the dataset's consistency and utility.

\subsection{Reasoning Examples}
\begin{figure*}[!h]
    \centering
    \includegraphics[width=\linewidth]{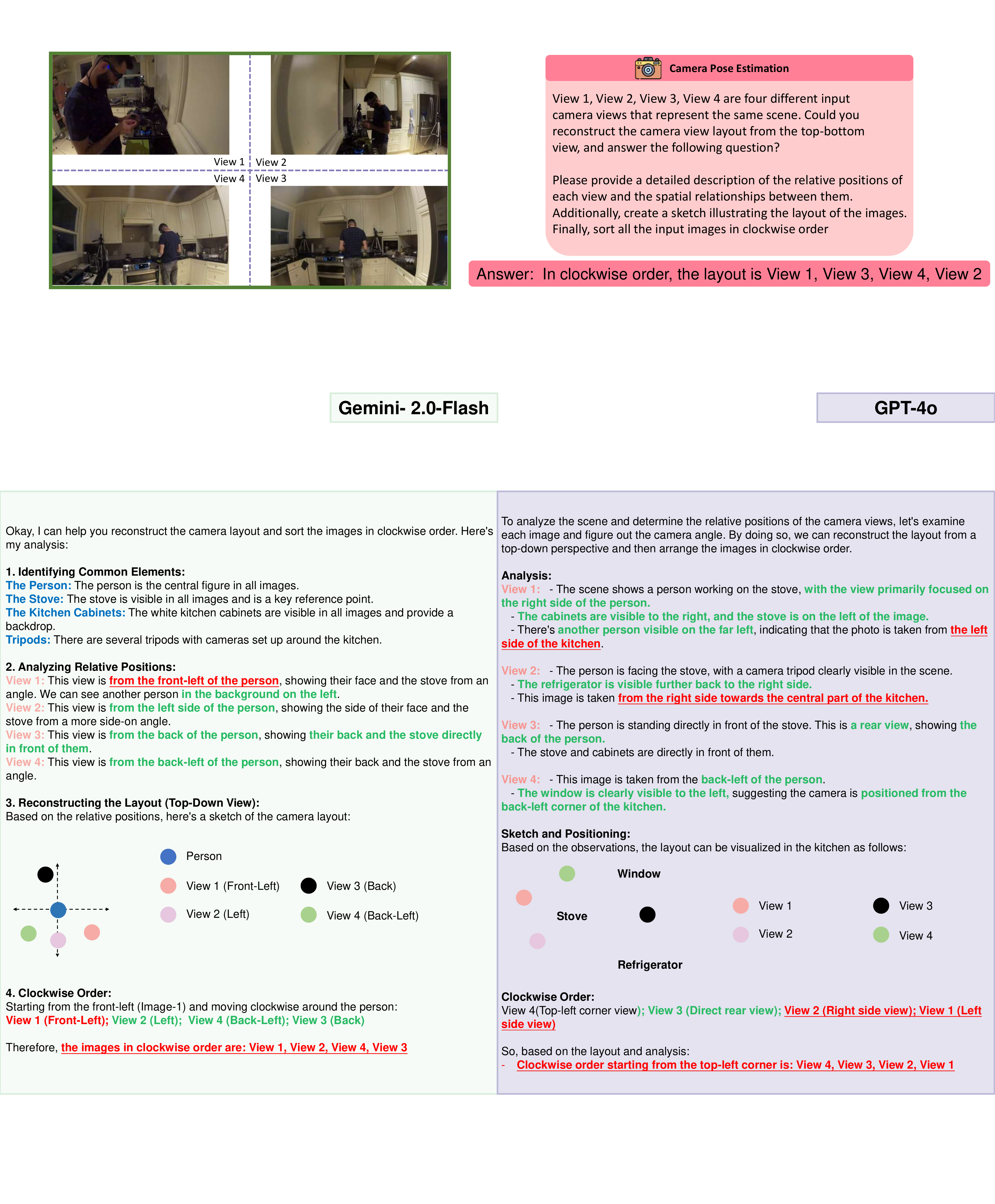}
    \caption{\textbf{Visualization of the reasoning process for two MLLMs (Part I).} In this case, both MLLMs choose incorrect options due to errors in their reasoning process.}
    \label{fig:reason_part1}
\end{figure*}

\begin{figure*}[!h]
    \centering
    \includegraphics[width=\linewidth]{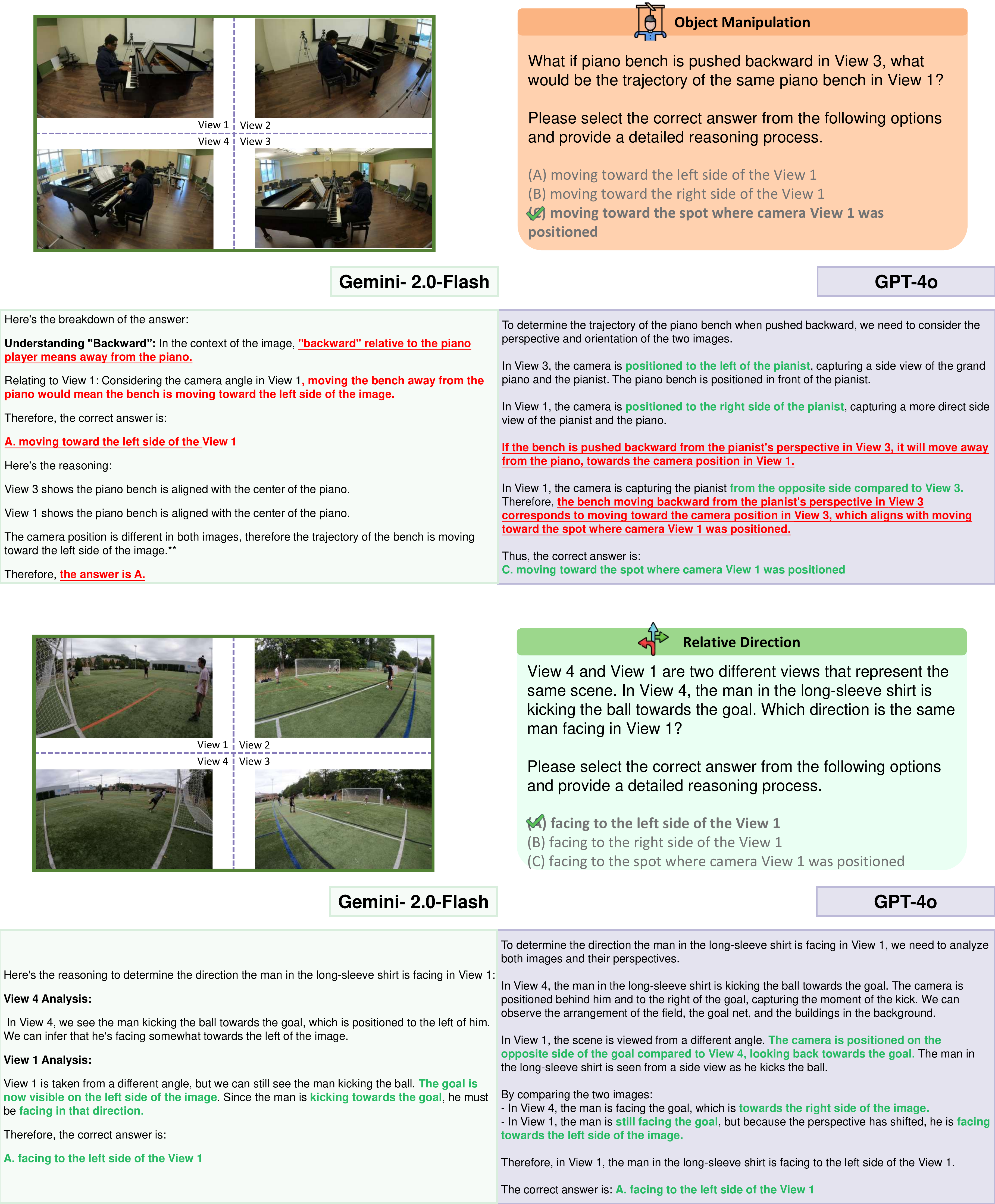}
    \caption{\textbf{Visualization of the reasoning process for two MLLMs (Part II).} In the above case, GPT-4o selects the correct option but contain errors in its reasoning process. In the case below, both GPT-4o and Gemini-2.0-Flash follow a correct reasoning process and ultimately select the right answer.}
    \label{fig:reason_part2}
\end{figure*}

Figures \ref{fig:reason_part1} and \ref{fig:reason_part2} depict three reasoning cases evaluated using GPT-4o~\cite{OpenAI2024gpt4o} and Gemini-2.0-Flash~\cite{team2023gemini}. We modify the post-prompt to instruct the models not only to generate answers but also to provide detailed reasoning. 
Our analysis reveals distinct reasoning patterns across the three evaluated cases. In case 1, both GPT-4o and Gemini-2.0-Flash select incorrect answers, indicating challenges in understanding the underlying spatial relationships. In case 2, while GPT-4o arrives at the correct answer, its reasoning process contains logical inconsistencies, suggesting that the model may have relied on heuristic shortcuts rather than fully comprehending the question. In contrast, Gemini-2.0-Flash fails to produce the correct response. While in case 3, both models correctly identify the answer, and their reasoning processes are coherent and logically sound, demonstrating their ability to follow step-by-step inference when the task aligns well with their learned knowledge.

\subsection{Human Evaluation Failure Cases}
\begin{figure*}[!h]
    \centering
    \includegraphics[width=\linewidth]{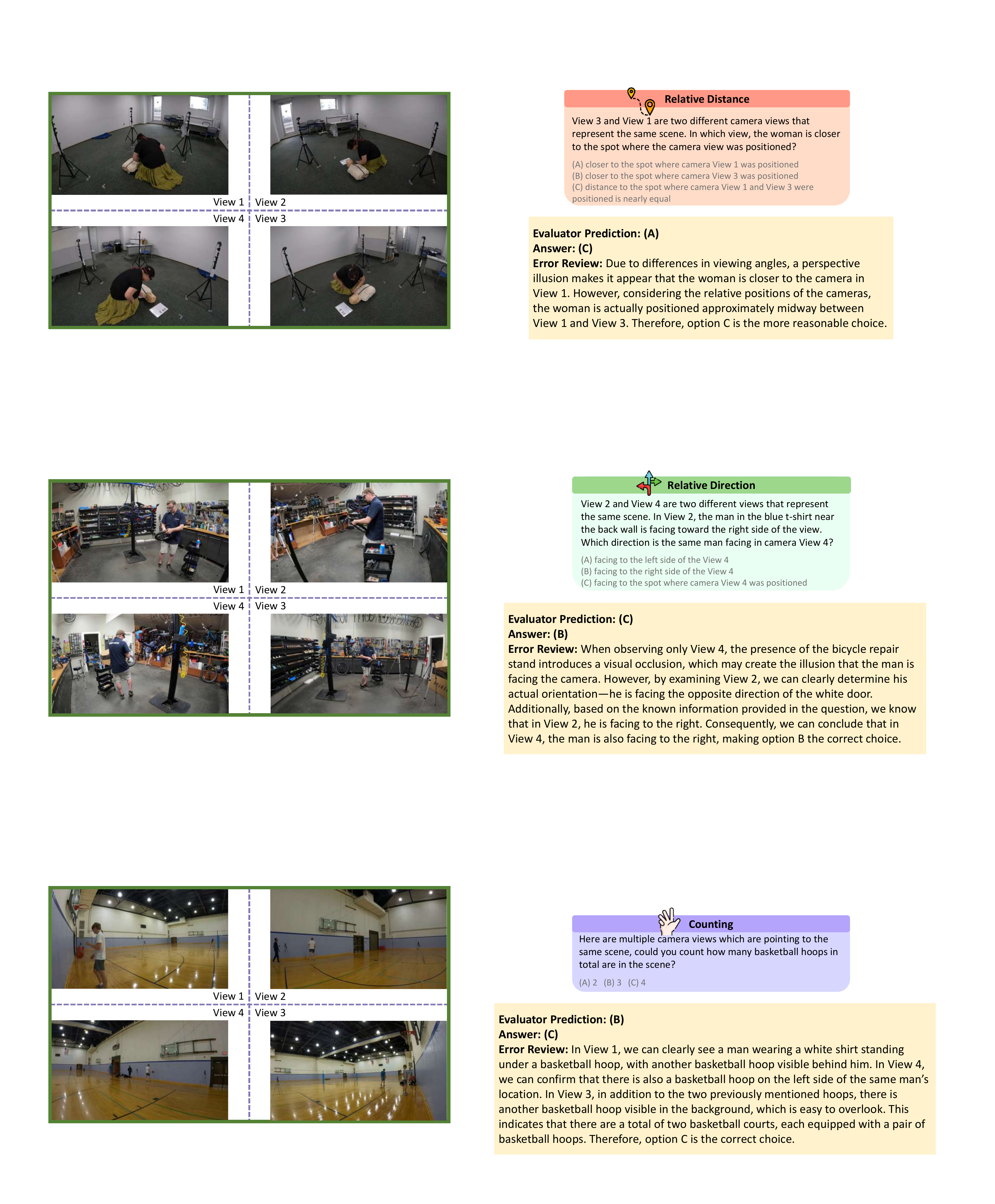}
    \caption{Questions that evaluators answered incorrectly, along with a detailed review of their reasoning for selecting the incorrect options.}
    \label{fig:human_failure}
\end{figure*}

\begin{figure*}[!h]
    \centering
    \includegraphics[width=2\linewidth]{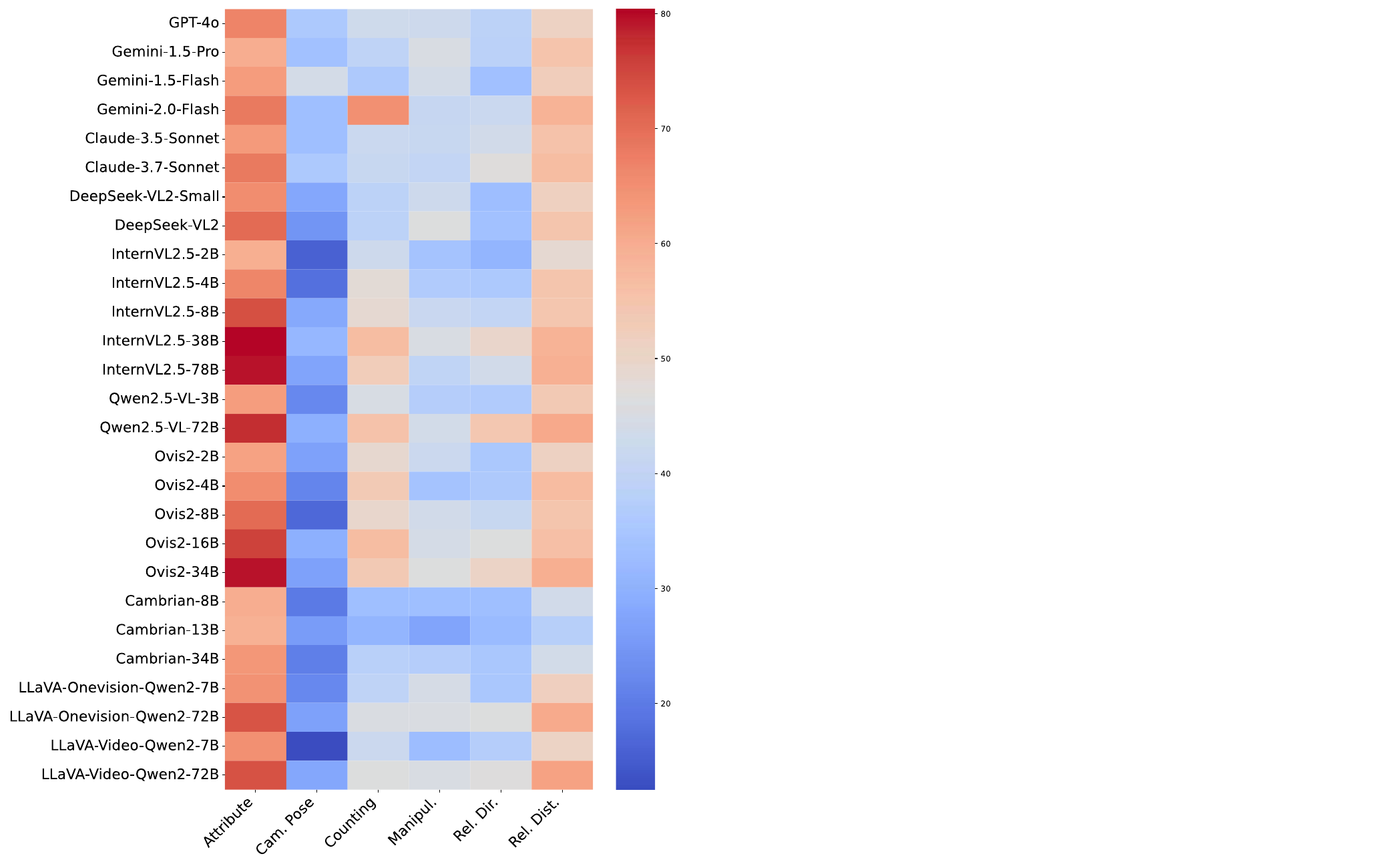}
    \caption{The visualization of all model performance across the 6 task categories in \textit{All-Angles Bench}.}
    \label{fig:confusion}
\end{figure*}

Figure \ref{fig:human_failure} presents three questions that human evaluators answered incorrectly. Analyzing these errors underscores the robustness of our annotation process while also highlighting the challenge and complexity of our benchmark. Additionally, the errors highlight the challenge and complexity of our benchmark, demonstrating its effectiveness in evaluating MLLMs' ability to understand spatial relationships in multi-view images. 

\subsection{Visualization Prompt}
Figure \ref{fig:figure8_prompt} displays the prompt used to visualize scene reconstruction and camera pose alignment, as detailed in our paper. This prompt allows us to assess how well GPT-4o and Gemini-2.0-Flash handle orientation-sensitive challenges, further validating their spatial reasoning capabilities.

\begin{figure*}[ht!]
    \captionsetup{type=table}
    \vspace{-0.4cm}
    \centering
    \fontsize{10.2pt}{10.0pt}\selectfont
    \setlength\tabcolsep{6pt}  
    \renewcommand{\arraystretch}{0.8}  
    \scalebox{1.0}{
    \begin{tabular}{r|c|cccccccc}
    & & 
    \rotatebox{30}{Attribute} &
    \rotatebox{30}{Cam. Pose} &
    \rotatebox{30}{Counting} & 
    \rotatebox{30}{Manipul.} &
    \rotatebox{30}{Rel. Dir.} &
    \rotatebox{30}{Rel. Dist.}      \\
    Methods & Avg. & \multicolumn{6}{c}{\cellcolor{pink!20}Multiple-Choice Answer} \\
    \hline
    Human Level  & 82.0  & 93.3  & 88.9  & 86.3  & 72.0  & 79.5  & 95.7 \\
    \rowcolor{green!10}
    \multicolumn{1}{l|}{\textcolor{black}{\textit{Closed-source Models}}}  & & & & & & &  \\
    GPT-4o & 52.4  & 66.7  & 16.7  & 52.9  & 40.0  & 53.8  & 63.8  \\ 
    Gemini-1.5-Pro & 50.8  & 60.0  & 22.2  & 35.3  & 44.0  & 51.3  & 76.6  \\ 
    Gemini-1.5-Flash & 50.0  & 55.6  & 22.2  & 27.5  & 50.0  & 56.4  & 74.5  \\ 
    Gemini-2.0-Flash & 58.4  & 62.2  & 38.9  & 64.7  & 48.0  & 56.4  & 68.1  \\ 
    Claude-3.5-Sonnet & 50.0  & 57.8  & 61.1  & 52.9  & 22.0  & 51.3  & 63.8  \\ 
    Claude-3.7-Sonnet & 52.8  & 60.0  & 38.9  & 37.3  & 38.0  & 56.4  & 80.9 \\
    \hline
    \rowcolor{green!10}
    \multicolumn{1}{l|}{\textcolor{black}{\textit{Open-source Models}}}  & & & & & & &  \\
    InternVL2.5-2B & 45.2  & 64.4  & 5.6  & 41.2  & 34.0  & 33.3  & 68.1  \\ 
    InternVL2.5-4B & 47.2  & 62.2  & 16.7  & 39.2  & 34.0  & 33.3  & 78.7  \\ 
    InternVL2.5-8B & 52.4  & 60.0  & 16.7  & 54.9  & 34.0  & 48.7  & 78.7  \\ 
    InternVL2.5-38B & 60.8  & 73.3  & 27.8  & 70.6  & 42.0  & 64.1  & 68.1  \\ 
    InternVL2.5-78B & 54.4  & 77.8  & 16.7  & 52.9  & 30.0  & 59.0  & 70.2  \\
    DeepSeek-VL2-Small & 48.0  & 62.2  & 38.9  & 45.1  & 32.0  & 35.9  & 68.1  \\ 
    DeepSeek-VL2 & 51.6  & 62.2  & 38.9  & 51.0  & 48.0  & 38.5  & 61.7  \\ 
    Qwen2.5-VL-3B & 52.4  & 68.9  & 22.2  & 43.1  & 42.0  & 46.2  & 74.5  \\ 
    Qwen2.5-VL-72B & 58.4  & 73.3  & 22.2  & 52.9  & 44.0  & 61.5  & 76.6  \\
    Ovis2-2B & 50.8  & 57.8  & 38.9  & 43.1  & 38.0  & 46.2  & 74.5  \\ 
    Ovis2-4B & 54.0  & 73.3  & 38.9  & 51.0  & 26.0  & 61.5  & 68.1  \\ 
    Ovis2-8B & 54.4  & 68.9  & 5.6  & 47.1  & 44.0  & 59.0  & 74.5  \\ 
    Ovis2-16B & 60.4  & 66.7  & 50.0  & 58.8  & 44.0  & 59.0  & 78.7  \\ 
    Ovis2-34B & 59.2  & 71.1  & 11.1  & 52.9  & 52.0  & 61.5  & 78.7  \\ 
    Cambrian-8B & 40.4  & 57.8  & 22.2  & 33.3  & 34.0  & 38.5  & 46.8  \\ 
    Cambrian-13B & 39.2  & 48.9  & 27.8  & 31.4  & 28.0  & 35.9  & 57.4  \\ 
    Cambrian-34B & 47.2  & 68.9  & 16.7  & 43.1  & 44.0  & 33.3  & 57.4  \\
    LLaVA-Onevision-Qwen2-7B & 53.6  & 64.4  & 16.7  & 43.1  & 44.0  & 53.8  & 78.7  \\ 
    LLaVA-Onevision-Qwen2-72B & 57.2  & 68.9  & 11.1  & 43.1  & 56.0  & 64.1  & 74.5  \\ 
    LLaVA-Video-Qwen2-7B & 46.8  & 62.2  & 5.6  & 43.1  & 26.0  & 53.8  & 68.1  \\ 
    LLaVA-Video-Qwen2-72B & 54.0  & 60.0  & 11.1  & 43.1  & 50.0  & 56.4  & 78.7  \\
    \hline
    \end{tabular}}
     \vspace{-0.2cm}
    \caption{Evaluation results for 27 MLLMs on 250 Q\&A tiny benchmark. }
    \label{tab:tiny_dataset_table}
    \vspace{-0.4cm}
\end{figure*}

\begin{figure*}
\begin{tcolorbox}[colback=black!5!white,colframe=black!75!black,title=Multiple-Choice Answer Extraction Prompt]

\texttt{Given a prediction for a multiple-choice question, directly extract the selected answer while skipping the reasoning process. If the prediction explicitly chooses option A, B, or C, return the corresponding letter. If the prediction does not specify a choice or indicates that none of the options are correct, return None.}

\end{tcolorbox}
\caption{The prompt used for extracting multiple-choice answers from predictions.}
\label{fig:fetching_prompts}
\end{figure*}

\begin{figure*}
\begin{tcolorbox}[colback=black!5!white,colframe=black!75!black,title=System Prompt]

\texttt{You are the expert for designing questions and answers for creating a benchmark for multi-camera view scenarios.}\\

\texttt{Here I will first give you several images from different camera views which are pointing to the same scene. 
Please answer the following questions based on the specific task description and example provide. 
Ensure the questions and answers are based on the objects and their relationships visible across the given camera views.}
\end{tcolorbox}
\caption{The system prompt used for generating five tasks with the MLLM.}
\label{fig:system_prompts}
\end{figure*}

\begin{figure*}
\begin{tcolorbox}[colback=black!5!white,colframe=black!75!black,title=Task Specific Prompt]

\texttt{\textbf{1. Counting:} Counting across multi-views} \\
\texttt{Task Description: Could you count the amount on specific object (e.g., people, chair, cup) in the scene based on all the input camera views?}\\
\texttt{Example:\\}
    \texttt{\textit{Question: Here are multiple camera views which are pointing to the same scene, could you count how many people in total are in the scene?}} \\
    \texttt{\textit{Option: (A) 5, (B) 3, (C) 1}} \\

\texttt{\textbf{2. Attribute Identification:} Attribute identification across multiple views} \\
\texttt{Task Description: Could you identify the same object / attribute across multiple camera views?} \\
\texttt{Example:\\}
    \texttt{\textit{Question: View 1 and View 2 are two different views that represent the same scene. In View 1, there is a person in fencing who is facing to the spot where camera View 1 was positioned. Could you identify the same person in View 2?}} \\
    \texttt{\textit{Option: (A) the person on the left side of View 2, (B) the person closer to the center of View 2, (C) the person sitting on the stairs}} \\ 

\texttt{\textbf{3. Relative Distance:} Object-camera relative distance}\\
\texttt{Task Description: Could you measure the relative distance between the seen object and camera view position in the cross view scenario?} \\
\texttt{Example:\\}
    \texttt{\textit{Question: View 1 and View 2 are two different camera views that represent the same scene. In which view, the basketball is closer to the spot where the camera view was positioned?}} \\
    \texttt{\textit{Option: (A) closer to the spot where camera View 1 was positioned, (B) closer to the spot where camera View 2 was positioned, (C) distance to the spot where camera View 1 and View 2 were positioned is equal}} \\

\texttt{\textbf{4. Relative Direction:} Object-camera relative direction}\\
\texttt{Task Description: Could you measure the relative direction between the seen object and camera view position in the cross view scenario?}\\
\texttt{Example:\\}
    \texttt{\textit{Question: View 1 and View 2 are two different camera views that represent the same scene. In View 1, the man in red t-shirt is facing to the spot where View 1 was positioned, which direction is the man facing in View 2?}}\\
    \texttt{\textit{Option: (A) facing to the left side of the view, (B) facing to the right side of the view, (C) facing to the spot where camera View 2 was positioned}}\\

\texttt{\textbf{5. Manipulation:} Relative object manipulation across views} \\
\texttt{Task Description: Could you manipulate the seen objects (e.g., what if the person is walking to the left of the view), and design the Q\&A based on the trajectory of object across views?}\\
\texttt{Example:\\}
    \texttt{\textit{Question: What if the person wearing the gray t-shirt is walking toward the spot where camera View 1 was positioned, what would be the trajectory of the same person in View 2?}}\\
    \texttt{\textit{Option: (A) walking toward the left side of View 2, (B) walking toward the right side of View 2, (C) remain in the same position}}

\end{tcolorbox}
\caption{The task-specific prompts used for generating five tasks with the MLLM.}
\label{fig:task_specific_prompts}
\end{figure*}

\begin{figure*}
\begin{tcolorbox}[colback=black!5!white,colframe=black!75!black,title=Camera Pose Estimation Question Template]

\texttt{\textit{Question: \\
View 1, View 2, View 3, View 4 are four different input
camera views that represent the same scene. Could you
reconstruct the camera view layout from the top-bottom
view, and answer the following question?}} \\

\texttt{\textit{Option: \\
(A) in clockwise order, the layout is View 1, View 2, View 3, View 4,\\
(B) in clockwise order, the layout is View 4, View 3, View 2, View 1, \\
(C) in clockwise order, the layout is View 2, View 3, View 1, View 4}}

\end{tcolorbox}
\caption{The question template designed for the Camera Pose Estimation task.}
\label{fig:cam_pose_template}
\end{figure*}


\begin{figure*}
\begin{tcolorbox}[colback=black!5!white,colframe=black!75!black]

\texttt{\textbf{[Task]}}

\texttt{These images capture the same scene. Your objective is to identify specific objects within each image, understand the spatial arrangement of the scene, and estimate the center point of each object, assuming the entire scene is represented by a 10x10 grid.}\\

\texttt{\textbf{[Rule]}}

\texttt{We provide the primary categories about in this scene: (e.g., human, lego, tv, sofa, stair...) Focus ONLY on these categories (ignore small object or categories). Estimate the center location of each instance within the provided categories, assuming the entire scene is represented by a 10x10 grid. If a category contains multiple instances, include all of them (e.g., Person A, Person B, Person C..). Each object’s estimated location should reflect its real position in the scene while preserving the relative spatial relationships. Combine and merge information from the images since they are pointing to the same scene, calibrating the object locations accordingly.}\\

\texttt{\textbf{[Output]}}

\texttt{Please generate a 10x10 grid visualization that includes: The predicted locations of each object (e.g., humans, basketball, etc.). The camera view position The camera’s viewpoint direction represented as an arrow indicating its view orientation The trajectory of the object in the question (e.g., object movement), indicating its motion path. Please also provide your answer and your reasons step by step in details.}

\end{tcolorbox}
\caption{Visualization prompt designed to evaluate MLLMs on orientation-sensitive challenges.}
\label{fig:figure8_prompt}
\end{figure*}
\clearpage

\end{document}